%% file: DLink.tex
\documentclass{article}

\PassOptionsToPackage{numbers, compress}{natbib}
\usepackage[preprint]{neurips_2026}

\usepackage[utf8]{inputenc}
\usepackage[T1]{fontenc}
\usepackage{courier}
\usepackage{hyperref}
\usepackage{url}
\usepackage{booktabs}
\usepackage{amsfonts}
\usepackage{nicefrac}
\usepackage{microtype}
\usepackage{xcolor}
\usepackage{graphicx}
\usepackage{amsmath}
\usepackage{amsthm}
\usepackage{amssymb}
\usepackage{algorithm}
\usepackage{algorithmic}
\usepackage{float}
\usepackage{capt-of}
\usepackage{enumitem}
\usepackage{makecell}
\usepackage{array}
\usepackage{multirow}

\title{DLink: Distilling Layer-wise and Dominant Knowledge from EEG Foundation Models}

\author{%
\textbf{Jingyuan Wang\textsuperscript{1},
Zhihao Jia\textsuperscript{1},
Chenyu Liu\textsuperscript{2},
Xinliang Zhou\textsuperscript{2},
Haoran Luo\textsuperscript{2},
Ziyu Jia\textsuperscript{3},}\\
\textbf{Yong Li\textsuperscript{4},
Fang Li\textsuperscript{2},
Junfeng Yao\textsuperscript{1,*},
Yi Ding\textsuperscript{2,*}}\\
\textsuperscript{1}Xiamen University \quad
\textsuperscript{2}Nanyang Technological University\\
\textsuperscript{3}Institute of Automation, Chinese Academy of Sciences \quad
\textsuperscript{4}Southeast University\\
\textsuperscript{*}Corresponding authors\\
\texttt{wjyuan@stu.xmu.edu.cn, jiazhihao@stu.xmu.edu.cn}\\
\texttt{chenyu003@e.ntu.edu.sg, xinliang001@e.ntu.edu.sg, haoran.luo@ieee.org}\\
\texttt{jia.ziyu@outlook.com, mysee1989@gmail.com, asfli@ntu.edu.sg}\\
\texttt{yao0010@xmu.edu.cn, ding.yi@ntu.edu.sg}
}

\begin{document}
\raggedbottom

\maketitle
\begin{abstract}
EEG foundation models (EFMs) achieve strong cross-subject and cross-task generalization through large-scale pretraining and downstream fine-tuning. Through empirical analysis, we observe that (i) task-adapted EFMs provide strong decoding performance but incur substantial overhead when retained as inference backbones, making knowledge distillation a natural route for optimizing compact students; and (ii) direct distillation from a fixed teacher representation underutilizes EFM knowledge, as task-discriminative information is distributed across intermediate layers rather than concentrated in the final layer. These observations motivate \textbf{DLink} (\textbf{D}istilling \textbf{L}ayer-w\textbf{i}se and Domi\textbf{n}ant \textbf{K}nowledge), a spectrally guided distillation framework with input-conditioned layer routing for transferring EFM knowledge into compact students.  DLink uses a lightweight router to aggregate teacher layers for each input, and aligns magnitude and phase spectra to mitigate compression-induced spectral distortion in learned representations. The routed teacher knowledge is internalized by a project-then-compress student; the teacher and router are used only during training. Experiments on four EEG benchmarks show that DLink improves matched compact students and remains competitive with lightweight baselines, narrowing the gap to fine-tuned EFMs while substantially reducing parameters, FLOPs, and CPU-only inference latency.
\end{abstract}

\section{Introduction}
Electroencephalogram (EEG) signals provide a non-invasive window into neural activity and have been used for brain--computer interfaces~\cite{EEG_data}, affective computing, cognitive assessment, and neurological disorder analysis~\cite{EEG2023,Sera,GIGN,Decoding_Attentive}. Recently, EEG foundation models (EFMs) have emerged as a promising paradigm for learning generalizable representations from large-scale heterogeneous EEG data~\cite{AdaBrain, EFM_worth}. By combining pretraining with downstream adaptation, these models have achieved strong cross-subject and cross-task performance, establishing EFMs as powerful backbones for EEG decoding.

\begin{figure}[!t] 
\centering 
\includegraphics[width=\columnwidth]{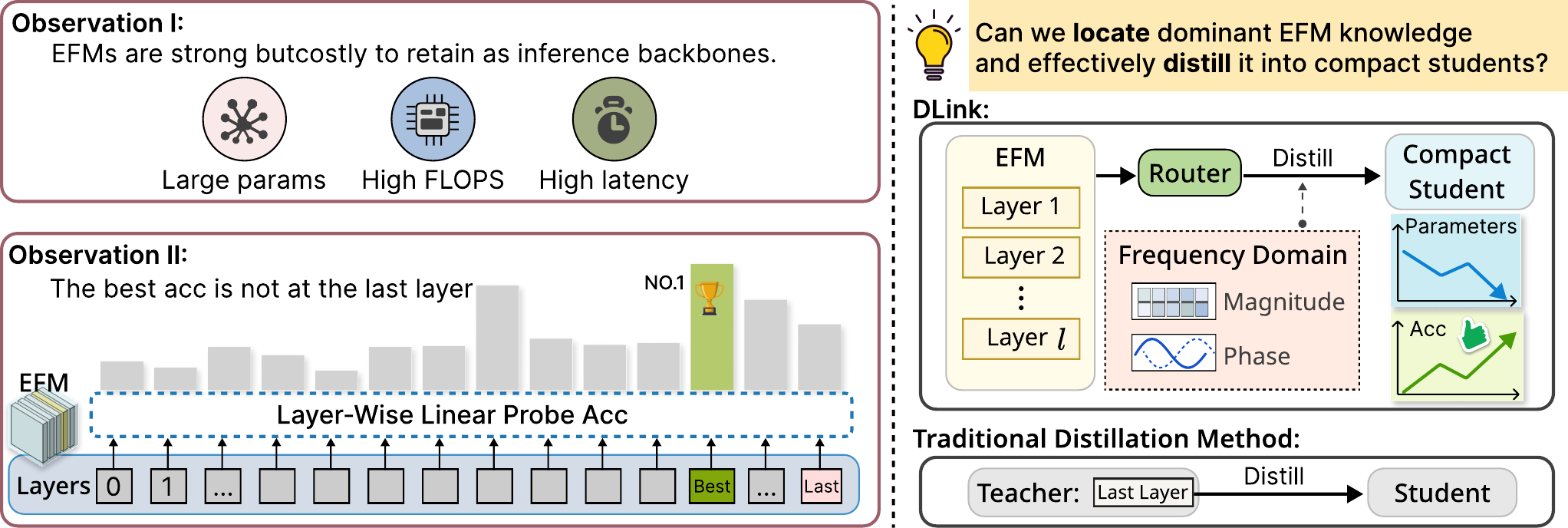} 
\caption{\textbf{Motivation and overview.} DLink is motivated by two observations. Compared with traditional final-layer distillation, DLink automatically aggregates layer-wise teacher knowledge and transfers it to a compact student through representation-level spectral consistency.}
\label{fig1} 
\end{figure}

Through empirical analysis, we first observe that 
\textbf{(I) task-adapted EFMs achieve strong decoding performance, but retaining their high-capacity backbones incurs substantial computational overhead}. 
Directly using an adapted EFM keeps the full teacher in the inference pipeline~\cite{EFM_capable}. 
Moreover, EFM hidden states are typically organized as large latent representations, so attaching task-specific heads often requires additional pooling, projection, or flattening~\cite{EFMsurvey}, further increasing parameter and computational cost~\cite{LUNA}. 
This cost motivates knowledge distillation as a natural route to transfer useful EFM knowledge into a compact student that is used alone after training.

Through layer-wise probing, we further observe that \textbf{(II) task-discriminative EFM information is often more prominent in intermediate layers than in the final representation}. Although conventional distillation commonly uses the final logits or hidden representation as the teacher signal~\cite{CMCRD}, Fig.~\ref{fig1} shows that intermediate layers can provide stronger discriminative features for downstream EEG decoding. This suggests that effective EFM distillation should aggregate dominant intermediate-layer knowledge rather than rely on a fixed final-layer source.

Even when informative intermediate layers are identified, \textbf{their high-dimensional representations remain difficult to internalize compactly}. Unlike final logits, intermediate EFM features retain high-dimensional teacher-side representations. Directly flattening these features would require an oversized task-specific classifier, whereas naive dimensionality reduction may discard discriminative structure during transfer~\cite{nipsBottleneck}. Moreover, compact students often rely on aggressive projection or downsampling to reduce feature size, which can distort the spectrum of the ordered temporal-latent representation entering the compression block. This suggests that EFM distillation should not only identify which teacher layers to transfer, but also regularize how high-dimensional teacher knowledge is mapped into compact student representations.

These observations and transfer constraints raise a central question:

\textbf{\emph{Can we locate dominant EFM knowledge and effectively distill it into compact students?}}

Addressing this question requires solving three technical challenges. \textbf{C1: Dominant knowledge localization and aggregation.} Because the useful teacher signal is not known a priori and may lie in intermediate layers, the distillation procedure should locate informative teacher-layer regions and aggregate them rather than rely on a fixed final representation. \textbf{C2: Compression-induced transfer loss.} The routed high-dimensional teacher representation must be transferred through projection and downsampling while preserving discriminative structure and limiting spectral distortion. \textbf{C3: Compact internalization.} The aggregated teacher knowledge must be absorbed into a lightweight inference path without a large classifier or teacher backbone.

To address these challenges, we propose \textbf{DLink} (\textbf{D}istilling \textbf{L}ayer-w\textbf{i}se and Domi\textbf{n}ant \textbf{K}nowledge), a spectrally guided distillation framework for transferring EFM knowledge into compact students. \textbf{For C1,} DLink introduces a lightweight layer router that learns soft weights over teacher layers from input-conditioned cues, allowing the student to receive supervision from dominant intermediate representations. \textbf{For C2,} DLink performs representation-level spectral distillation by aligning magnitude and phase spectra between the routed teacher representation and the student feature, which regularizes compact transfer and mitigates spectral distortion. \textbf{For C3,} DLink uses a compact project-then-compress student: it first maps the input into a feature space aligned with teacher features and then applies structured spatio-temporal compression before task prediction. Importantly, the teacher and router are discarded after training, leaving only the compact student for inference.

We evaluate DLink on four EEG benchmarks covering diverse downstream decoding tasks. Experiments show that DLink improves matched compact students under ACC-B across all benchmarks and remains competitive with lightweight EEG baselines, while narrowing the gap to fine-tuned EFMs with substantially lower inference cost. Further analyses verify the roles of learned layer aggregation and spectral distillation, and show that DLink adds little training overhead. Together, these results demonstrate that DLink provides a practical framework for distilling layer-wise dominant knowledge from EFMs into compact EEG models.

\section{Preliminaries and Problem Formulation}

\subsection{EEG Foundation Models}

Let $X \in \mathcal{X}$ denote an EEG sample and $y \in \mathcal{Y}$ its corresponding label. 
An EFM teacher $\mathcal{T}$ represents $X$ as latent units and produces $L$ layer-wise hidden representations:
\begin{equation}
    f_T^{(l)} = \mathcal{T}^{(l)}(X) \in \mathbb{R}^{N_l \times d_l}, 
    \quad l=1,\dots,L,
\end{equation}
where $N_l$ and $d_l$ denote the latent resolution and feature dimension. 
For methods that require output logits, the final representation can be passed to a prediction head; DLink itself uses only hidden representations from the teacher. 
Representative EFMs, such as LaBraM, CBraMod, and EEG-DINO, instantiate this form with different encoders~\cite{LarBraM,CBraMod,EEG_DINO}. 
This notation emphasizes that $f_T^{(L)}$ is only one candidate source within the hierarchy $\{f_T^{(l)}\}_{l=1}^{L}$.

In our implementation, DLink is instantiated separately for each teacher EFM. The router aggregates hidden states only within a single teacher, using the routed hidden-state list exposed at a common latent layout and hidden width after that teacher's tokenization. Accordingly, in the method section $L$ denotes this routed layer list, and we write $f_T^{(l)}\in\mathbb{R}^{N\times d}$ for all routed layers.

\subsection{Direct Teacher--Student Distillation}

Given a compact student $\mathcal{S}$, teacher--student distillation aims to transfer knowledge from the high-capacity teacher $\mathcal{T}$ into $\mathcal{S}$ while preserving downstream prediction performance~\cite{SFTN,KDsurvey,DistiLLM}. 
Let $f_S=\mathcal{S}_{\mathrm{feat}}(X)$ and $\hat{y}_S=h_S(f_S)$ denote the student feature and prediction. 
Direct distillation typically uses the teacher's final logits or final hidden representation as the supervision target:
\begin{equation}
    \mathcal{L}_{\mathrm{direct}}
    =
    \mathcal{L}_{\mathrm{cls}}(\hat{y}_S,y)
    +
    \lambda \mathcal{L}_{\mathrm{KD}},
    \quad
    \mathcal{L}_{\mathrm{KD}}
    =
    \begin{cases}
    \tau^2 D_{\mathrm{KL}}\!\left(\sigma(z_T/\tau)\,\|\,\sigma(z_S/\tau)\right), & \text{logit KD},\\
    d\left(\tilde{f}_S, f_T^{(L)}\right), & \text{feature KD},
    \end{cases}
\end{equation}
where $z_T$ and $z_S$ denote teacher- and student-side logits when a logit-KD baseline is instantiated, $\tilde{f}_S$ is the dimension-matched student feature used by the feature-KD baseline, $\sigma(\cdot)$ is softmax, $\tau$ is the temperature, and $d(\cdot,\cdot)$ is a feature discrepancy measure. In DLink, the teacher is used as a frozen representation encoder and no teacher-side classifier is required.
This fixed-target formulation is simple, but it can underuse EFMs whose task-relevant information is distributed across intermediate layers~\cite{CMCRD,Less_is_more,FMsurvey}. 
It also leaves open how a compact student should inherit high-dimensional teacher knowledge without losing discriminative structure during compression.

\subsection{Problem Statement: Dominant Knowledge Distillation under Compression}
\label{subsec:problem_statement}

We formulate EFM distillation as a problem of dominant knowledge distillation under compression. 
Given a teacher EFM $\mathcal{T}$ with layer-wise representations $\{f_T^{(l)}\}_{l=1}^{L}$, the goal is to train a compact student $\mathcal{S}$ that is used alone at inference time while retaining the task-relevant knowledge of the teacher. 
Instead of prescribing a fixed teacher layer, we introduce a dominant knowledge operator $\mathcal{G}$:
\begin{equation}
    f_T^{*}
    =
    \mathcal{G}
    \left(
    \{f_T^{(l)}\}_{l=1}^{L}, X
    \right),
\end{equation}
where $f_T^{*}$ denotes the aggregated teacher knowledge used to supervise the student. 
The student, however, cannot simply copy this representation: $f_T^{*}$ may be high-dimensional and teacher-specific, whereas the student must encode it into a much smaller carrier. 
We refer to the task-relevant structure lost in this high-to-low-dimensional transfer as \emph{compression-induced transfer loss}. 
The general objective is therefore
\begin{equation}
    \min_{\theta_S, \theta_G}
    \;
    \mathbb{E}_{(X,y)}
    \left[
    \mathcal{L}_{\mathrm{cls}}(\hat{y}_S,y)
    +
    \lambda
    \mathcal{D}_{\mathrm{comp}}
    \left(
    \psi_S(X), f_T^{*}
    \right)
    \right],
    \quad
    \mathrm{s.t.}\;\; B(\mathcal{S}) \ll B(\mathcal{T}),
\end{equation}
where $\psi_S(X)$ denotes the student representation used for distillation, $\mathcal{D}_{\mathrm{comp}}(\cdot,\cdot)$ is the discrepancy used for compact transfer, and $B(\cdot)$ denotes an inference budget such as parameters or FLOPs. 

This formulation gives three requirements: automatically aggregating dominant teacher layers (C1), preserving teacher-induced structure during compact transfer (C2), and keeping the inference path teacher-free and lightweight (C3).

\begin{figure*}[!t]
    \centering 
\includegraphics[width=0.99\linewidth]{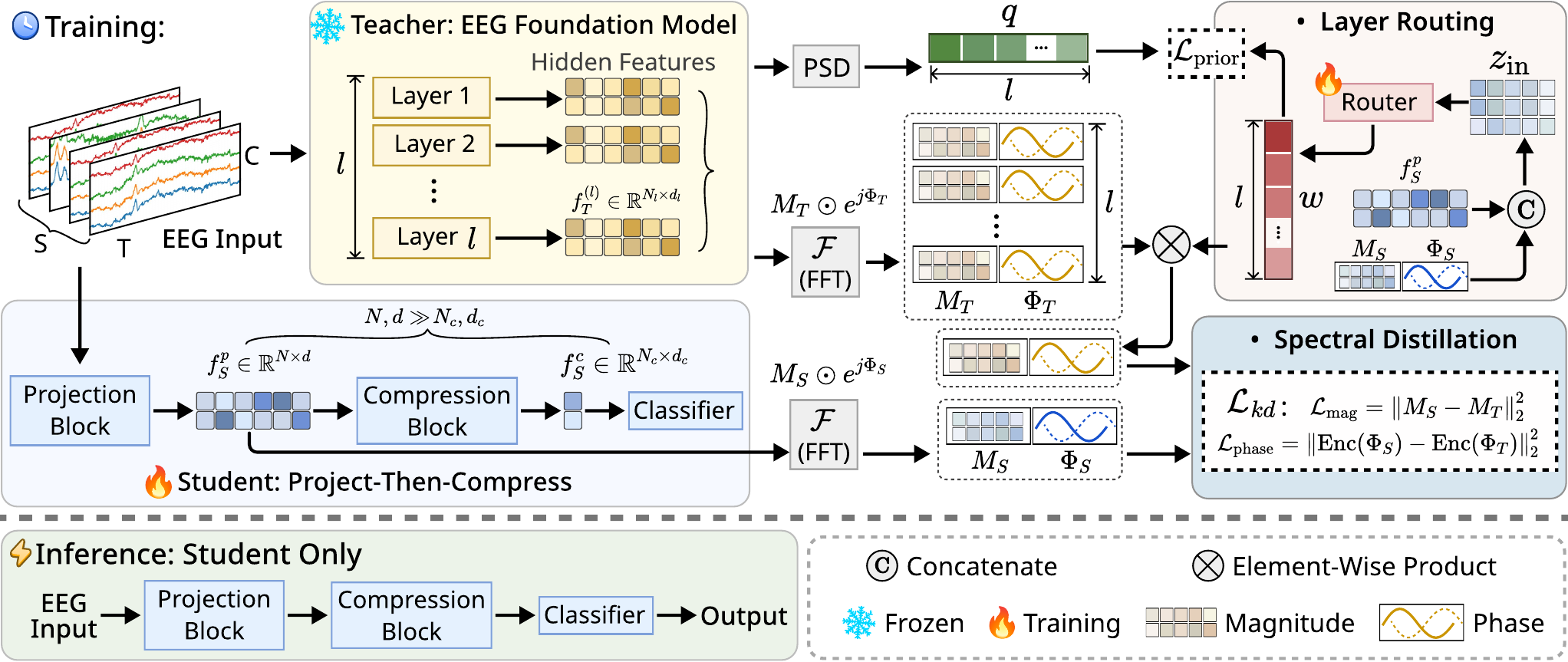}
\caption{\textbf{Overview of the DLink pipeline.} DLink transfers knowledge from a frozen EEG foundation-model teacher to a compact student. During training, an input-conditioned layer router softly aggregates dominant teacher-layer representations, and spectral distillation aligns the resulting routed teacher representation with the student feature in learned representation spectra. After distillation, both the teacher and router are discarded, leaving only the compact student for inference.}
    \label{fig:dlink_overview}
\end{figure*}

\section{Method: DLink}
\label{sec:methodology}

\subsection{Framework Overview}

Given a frozen teacher EFM $\mathcal{T}$ and a student $\mathcal{S}$, DLink distills layer-wise teacher representations into the student while discarding the teacher at inference. During DLink optimization, the teacher encoder is fixed and no teacher-side classifier is used. As shown in Fig.~\ref{fig:dlink_overview}, DLink combines a layer router, representation-level spectral distillation, and a compact project-then-compress student.
The student exposes two internal representations:
\begin{equation}
    f_S^{p} = P_{\theta_S}(X) \in \mathbb{R}^{N \times d}, 
    \qquad 
    f_S^{c} = C_{\theta_S}(f_S^{p}) \in \mathbb{R}^{N_c \times d_c},
\end{equation}
where $P_{\theta_S}(\cdot)$ is the projection block for teacher--student alignment, $C_{\theta_S}(\cdot)$ is the compression block, and $N_c d_c \ll N d$. The target shape $(N,d)$ is teacher-specific and is matched to the common shape of the routed hidden layers for the current teacher. The representation $f_S^{p}$ is used for distillation, while $f_S^{c}$ is used for compact prediction.

\subsection{Dominant Intermediate Layer Aggregation}
\label{subsec:router}

DLink avoids assuming that the final teacher layer is always the most informative supervision source. Instead, it learns a lightweight layer router that assigns soft importance weights to teacher layers from student-side cues:
\begin{equation}
    z_{\mathrm{in}} = 
    \left[
    \mathrm{Pool}(f_S^{p}) \; \Vert \; 
    \mathrm{Pool}\left(|\mathcal{F}_N(f_S^{p})|\right)
    \right],
\end{equation}
where $\mathcal{F}_N(\cdot)$ denotes a one-dimensional Fourier transform along the ordered temporal-latent axis $N$, $|\mathcal{F}_N(\cdot)|$ denotes the corresponding representation spectrum, $\mathrm{Pool}(\cdot)$ denotes lightweight pooling over the resulting feature map, and $\Vert$ denotes concatenation. The router maps $z_{\mathrm{in}}$ to layer-wise logits:
\begin{equation}
    a = R_{\theta_R}(z_{\mathrm{in}}) \in \mathbb{R}^{L},
\end{equation}
which are normalized by a temperature-scaled softmax:
\begin{equation}
    w_l = 
    \frac{\exp(a_l/\tau)}
    {\sum_{k=1}^{L}\exp(a_k/\tau)},
    \qquad l=1,\dots,L.
\end{equation}
The resulting vector $w=[w_1,\dots,w_L]$ defines an input-conditioned distribution over the routed teacher layers. The routed teacher representation is obtained by weighted aggregation:
\begin{equation}
    f_T^{*} = \sum_{l=1}^{L} w_l f_T^{(l)}.
\end{equation}

To stabilize routing, we further introduce a representation-level spectral-concentration prior over teacher layers. For each input and teacher representation $f_T^{(l)}$, we compute aggregated spectral energy over representation-frequency bins induced by the temporal-latent axis:
\begin{equation}
    e_l(\omega)
    =
    \left\|
    \mathcal{F}_N(f_T^{(l)})(\omega)
    \right\|_2^2,
    \qquad
    p_l(\omega)
    =
    \frac{e_l(\omega)}
    {\sum_{\omega'\in\Omega} e_l(\omega')+\epsilon},
\end{equation}
where $\Omega$ indexes representation-frequency components along the temporal-latent axis. This prior is computed from learned teacher representations and should not be interpreted as raw-EEG spectral supervision. We reduce this distribution to a scalar layer score by its spectral concentration:
\begin{equation}
    s_l = \sum_{\omega\in\Omega} p_l(\omega)^2.
\end{equation}
For each input, the layer scores are normalized into a soft guidance distribution with guidance temperature $\tau_q$:
\begin{equation}
    q_l =
    \frac{\exp(s_l/\tau_q)}
    {\sum_{k=1}^{L}\exp(s_k/\tau_q)}.
\end{equation}
The router is regularized by
\begin{equation}
    \mathcal{L}_{\mathrm{prior}}
    =
    D_{\mathrm{KL}}(w \,\|\, q).
\end{equation}
This guidance is not a hard layer label. It provides a representation-level spectral-concentration cue, while the final routing behavior is jointly shaped by the distillation and task objectives.

\subsection{Representation-Spectrum Consistency for Compact Transfer}
\label{distill_mechanism}

DLink transfers the routed teacher representation to the student through representation-spectrum consistency. For the teacher--student feature pairs used by DLink, we write $f\in\mathbb{R}^{N\times d}$ as a notational view: $N$ denotes the ordered temporal-latent axis used for spectral alignment, while $d$ denotes the remaining feature indices. We apply a one-dimensional Fourier transform $\mathcal{F}_N$ along the $N$ axis, independently over the remaining feature indices. Thus, the spectral bins describe coarse-to-fine variation across adjacent temporal latent units in the learned representation; they are not raw EEG frequencies in Hz. This axis is meaningful for compact transfer because the compression block reduces temporal latent resolution through projection and spatio-temporal downsampling. We therefore apply spectral distillation before compression as a feature-level regularizer.

A direct feature-distillation baseline would align the same routed teacher representation by pointwise matching, which constrains Euclidean discrepancy but does not explicitly regulate the temporal-latent spectrum entering the compression block. DLink instead constrains the magnitude and phase spectra of $f_S^{p}$ to follow those of $f_T^{*}$.

Given the student projection feature $f_S^{p}$ and the routed teacher representation $f_T^{*}$, we compute their spectral representations:
\begin{equation}
    \mathcal{F}_N(f_S^{p}) = M_S \odot e^{j\Phi_S},
    \qquad
    \mathcal{F}_N(f_T^{*}) = M_T \odot e^{j\Phi_T},
\end{equation}
where $M_S,M_T$ are magnitude spectra and $\Phi_S,\Phi_T$ are phase spectra. The magnitude term aligns representation-level spectral magnitudes:
\begin{equation}
    \mathcal{L}_{\mathrm{mag}}
    =
    \left\| M_S - M_T \right\|_F^2.
\end{equation}
Directly matching raw phase values is unstable due to the $2\pi$ periodicity. Following a continuous phase encoding, we represent phase by sine and cosine components:
\begin{equation}
    \mathrm{Enc}(\Phi) = [\cos \Phi, \sin \Phi].
\end{equation}
The phase consistency term is then defined as
\begin{equation}
    \mathcal{L}_{\mathrm{phase}}
    =
    \left\|
    \mathrm{Enc}(\Phi_S) -
    \mathrm{Enc}(\Phi_T)
    \right\|_F^2.
\end{equation}
The spectral distillation objective is
\begin{equation}
    \mathcal{L}_{\mathrm{kd}}
    =
    \mathcal{L}_{\mathrm{mag}}
    +
    \mathcal{L}_{\mathrm{phase}}.
\end{equation}

\subsection{Compact Project-then-Compress Student}

The student is designed as a compact carrier for internalizing routed teacher knowledge. It follows a project-then-compress principle. The CNN--Transformer projection block first produces $f_S^{p}$ for spectral distillation against $f_T^{*}$. A CNN-based compression block then applies spatio-temporal downsampling to obtain the compact feature $f_S^{c}$, which is fed into a compact MLP head for prediction, i.e., $\hat{y}_S=h_{\theta_S}(f_S^{c})$.
This design separates knowledge alignment from compact prediction. The CNN--Transformer projection block provides sufficient capacity to absorb teacher-side supervision, while CNN downsampling prevents the MLP classifier from directly operating on oversized teacher-like features. Detailed layer configurations and student size variants are provided in Appendix~\ref{app:student_structure}.

\subsection{Training Objective and Inference Path}
\label{subsec:optimization}

The final training objective combines task supervision, spectral distillation, and router guidance:
\begin{equation}
\label{eq: total_loss}
    \mathcal{L}_{\mathrm{total}}
    =
    \mathcal{L}_{\mathrm{cls}}(\hat{y}_S,y)
    +
    \lambda \mathcal{L}_{\mathrm{kd}}
    +
    \mathcal{L}_{\mathrm{prior}}.
\end{equation}
During training, the teacher EFM $\mathcal{T}$ is frozen and provides only layer-wise representations for DLink. The student $\mathcal{S}$ and router $R_{\theta_R}$ are optimized jointly.
Inference is performed only by the student.


\section{Experiments}

We evaluate DLink from four aspects: compact-student accuracy, learned layer aggregation, spectral transfer behavior, and inference efficiency.

\subsection{Experimental Setup}

\textbf{Datasets and metrics.}
We evaluate DLink on four EEG benchmarks: \textbf{FACED}~\cite{FACED} for 9-class emotion recognition, \textbf{Mumtaz2016}~\cite{Mumtaz2016} for binary depression diagnosis, \textbf{PhysioNet-MI}~\cite{PhysioNet2000,PhysioNet2004} for 4-class motor imagery, and \textbf{SHU-MI}~\cite{SHU} for binary motor imagery. Following the CBraMod downstream protocol~\cite{CBraMod}, EEG signals are resampled to 200 Hz and represented as 1-second patches.
Dataset statistics, split protocols, leakage-control details, and metric definitions are provided in the Appendix for full experimental context.

\textbf{Baselines.}
We compare DLink with three groups of baselines. First, we include lightweight EEG models, including EEGNet~\cite{EEGNet}, EEGConformer~\cite{conformer}, and EEGDeformer~\cite{Deformer}. Second, we report fully fine-tuned EFMs, including CBraMod~\cite{CBraMod}, LaBraM-Base~\cite{LarBraM}, and EEG-DINO~\cite{EEG_DINO}, as high-capacity teacher references rather than lightweight competitors. Third, we compare with direct teacher--student distillation baselines, including FitNets~\cite{FitNets} for feature-level distillation and Logit-std~\cite{Logit_std} for logit-level distillation. All distillation baselines use the same teacher--student configuration as DLink for fair comparison.

\textbf{Implementation.}
We evaluate two student scales, Stu-S and Stu-M; DLink-S/M denote applying DLink to them. For Eq.~\ref{eq: total_loss}, the only tuned loss weight is the spectral coefficient $\lambda$, with dataset settings: FACED (LR $2\times10^{-3}$, $\lambda=0.5$), Mumtaz2016 (LR $8\times10^{-3}$, $\lambda=0.2$), PhysioNet-MI (LR $2\times10^{-3}$, $\lambda=0.8$), and SHU-MI (LR $5\times10^{-4}$, $\lambda=0.2$). For each run, we select the checkpoint with the highest validation accuracy and report the test result averaged over five seeds. Optimization details and sensitivity analyses are provided in Appendix~\ref{app:implementation_details} and Appendix~\ref{app:hyperparameter_sensitivity}.

\begin{table*}[t]
\centering
\caption{Main results across four datasets (\textbf{bold}/\underline{underline}: top two results among lightweight models only; fully fine-tuned teachers are reported as high-capacity references and are not included in this ranking). For fairness, all teachers and \textbf{Stu-M} share a consistent 200-unit hidden layer in their classification MLP heads.}
\label{tab:main_results}
\vspace{0.4em}
\setlength{\tabcolsep}{8pt}
\resizebox{1\textwidth}{!}{
\begin{tabular}{ll|ccc|ccc}
\toprule
\multirow{2}{*}{\textbf{Category}} & \multirow{2}{*}{\textbf{Model}} & \multicolumn{3}{c|}{\textbf{FACED (9-class)}} & \multicolumn{3}{c}{\textbf{Mumtaz2016 (2-class)}} \\ 
 & & \textbf{ACC-B} & \textbf{F1-W} & \textbf{Kappa} & \textbf{ACC-B} & \textbf{AUROC} & \textbf{AUC-PR} \\ \midrule
\multirow{5}{*}{\makecell{Small\\Models}} 
& EEGNet & 0.2281 {\scriptsize $\pm$ 0.0265} & 0.1771 {\scriptsize $\pm$ 0.0353} & 0.1294 {\scriptsize $\pm$ 0.0306} & 0.8958 {\scriptsize $\pm$ 0.0056} & 0.9616 {\scriptsize $\pm$ 0.0128} & 0.9678 {\scriptsize $\pm$ 0.0073} \\
& EEGConformer & 0.4324 {\scriptsize $\pm$ 0.0188} & 0.4405 {\scriptsize $\pm$ 0.0188} & 0.3772 {\scriptsize $\pm$ 0.0086} & 0.8923 {\scriptsize $\pm$ 0.0075} & 0.9502 {\scriptsize $\pm$ 0.0060} & 0.9526 {\scriptsize $\pm$ 0.0080} \\
& EEGDeformer & 0.3440 {\scriptsize $\pm$ 0.0126} & 0.3408 {\scriptsize $\pm$ 0.0145} & 0.2592 {\scriptsize $\pm$ 0.0123} & 0.9003 {\scriptsize $\pm$ 0.0247} & \textbf{0.9760 {\scriptsize $\pm$ 0.0137}} & \textbf{0.9806 {\scriptsize $\pm$ 0.0156}} \\
& Stu-S & 0.4118 {\scriptsize $\pm$ 0.0206} & 0.4109 {\scriptsize $\pm$ 0.0192} & 0.3355 {\scriptsize $\pm$ 0.0225} & 0.8853 {\scriptsize $\pm$ 0.0175} & 0.9634 {\scriptsize $\pm$ 0.0103} & 0.9692 {\scriptsize $\pm$ 0.0089} \\
& Stu-M & 0.5038 {\scriptsize $\pm$ 0.0172} & 0.5063 {\scriptsize $\pm$ 0.0158} & 0.4390 {\scriptsize $\pm$ 0.0190} & 0.8975 {\scriptsize $\pm$ 0.0171} & 0.9648 {\scriptsize $\pm$ 0.0078} & 0.9702 {\scriptsize $\pm$ 0.0069} \\ \midrule
\multirow{3}{*}{\makecell{Fine-tuned\\Teachers}} 
& CBraMod & 0.4915 {\scriptsize $\pm$ 0.0050} & 0.4888 {\scriptsize $\pm$ 0.0050} & 0.4240 {\scriptsize $\pm$ 0.0059} & 0.9011 {\scriptsize $\pm$ 0.0076} & 0.9753 {\scriptsize $\pm$ 0.0070} & 0.9785 {\scriptsize $\pm$ 0.0048} \\
& LaBraM & 0.4793 {\scriptsize $\pm$ 0.0032} & 0.4805 {\scriptsize $\pm$ 0.0033} & 0.4116 {\scriptsize $\pm$ 0.0034} & 0.8968 {\scriptsize $\pm$ 0.0018} & 0.9706 {\scriptsize $\pm$ 0.0047} & 0.9732 {\scriptsize $\pm$ 0.0038} \\
& EEG-DINO & 0.4461 {\scriptsize $\pm$ 0.0071} & 0.4451 {\scriptsize $\pm$ 0.0059} & 0.3735 {\scriptsize $\pm$ 0.0080} & 0.8851 {\scriptsize $\pm$ 0.0041} & 0.9735 {\scriptsize $\pm$ 0.0042} & 0.9739 {\scriptsize $\pm$ 0.0038} \\ \midrule
\multirow{3}{*}{\makecell{KD\\(Stu-S)}} 
& FitNets & 0.4239 {\scriptsize $\pm$ 0.0068} & 0.4217 {\scriptsize $\pm$ 0.0067} & 0.3486 {\scriptsize $\pm$ 0.0083} & 0.8801 {\scriptsize $\pm$ 0.0042} & 0.9631 {\scriptsize $\pm$ 0.0069} & 0.9697 {\scriptsize $\pm$ 0.0056} \\
& Logit-std & 0.4125 {\scriptsize $\pm$ 0.0125} & 0.4117 {\scriptsize $\pm$ 0.0127} & 0.3367 {\scriptsize $\pm$ 0.0141} & 0.8491 {\scriptsize $\pm$ 0.0234} & 0.9325 {\scriptsize $\pm$ 0.0161} & 0.9449 {\scriptsize $\pm$ 0.0130} \\
& \textbf{DLink (Ours)} & 0.4330 {\scriptsize $\pm$ 0.0132} & 0.4251 {\scriptsize $\pm$ 0.0107} & 0.3548 {\scriptsize $\pm$ 0.0113} & \underline{0.9010 {\scriptsize $\pm$ 0.0045}} & 0.9612 {\scriptsize $\pm$ 0.0044} & 0.9683 {\scriptsize $\pm$ 0.0034} \\ \midrule
\multirow{3}{*}{\makecell{KD\\(Stu-M)}} 
& FitNets & 0.4858 {\scriptsize $\pm$ 0.0259} & 0.4889 {\scriptsize $\pm$ 0.0234} & 0.4191 {\scriptsize $\pm$ 0.0281} & 0.8963 {\scriptsize $\pm$ 0.0131} & 0.9674 {\scriptsize $\pm$ 0.0080} & 0.9731 {\scriptsize $\pm$ 0.0064} \\
& Logit-std & \underline{0.5122 {\scriptsize $\pm$ 0.0077}} & \underline{0.5117 {\scriptsize $\pm$ 0.0071}} & \underline{0.4483 {\scriptsize $\pm$ 0.0078}} & 0.8937 {\scriptsize $\pm$ 0.0230} & 0.9464 {\scriptsize $\pm$ 0.0377} & 0.9633 {\scriptsize $\pm$ 0.0212} \\
& \textbf{DLink (Ours)} & \textbf{0.5221 {\scriptsize $\pm$ 0.0052}} & \textbf{0.5202 {\scriptsize $\pm$ 0.0074}} & \textbf{0.4581 {\scriptsize $\pm$ 0.0066}} & \textbf{0.9024 {\scriptsize $\pm$ 0.0082}} & \underline{0.9695 {\scriptsize $\pm$ 0.0074}} & \underline{0.9738 {\scriptsize $\pm$ 0.0060}} \\ \midrule \midrule
\multirow{2}{*}{\textbf{Category}} & \multirow{2}{*}{\textbf{Model}} & \multicolumn{3}{c|}{\textbf{PhysioNet-MI (4-class)}} & \multicolumn{3}{c}{\textbf{SHU (2-class)}} \\ 
 & & \textbf{ACC-B} & \textbf{F1-W} & \textbf{Kappa} & \textbf{ACC-B} & \textbf{AUROC} & \textbf{AUC-PR} \\ \midrule
\multirow{5}{*}{\makecell{Small\\Models}} 
& EEGNet & 0.5890 {\scriptsize $\pm$ 0.0062} & 0.5968 {\scriptsize $\pm$ 0.0133} & \textbf{0.4853 {\scriptsize $\pm$ 0.0076}} & 0.5714 {\scriptsize $\pm$ 0.0221} & 0.6484 {\scriptsize $\pm$ 0.0125} & 0.6305 {\scriptsize $\pm$ 0.0115} \\
& EEGConformer & 0.5879 {\scriptsize $\pm$ 0.0057} & 0.5878 {\scriptsize $\pm$ 0.0056} & 0.4504 {\scriptsize $\pm$ 0.0077} & 0.6014 {\scriptsize $\pm$ 0.0236} & 0.6651 {\scriptsize $\pm$ 0.0222} & 0.6463 {\scriptsize $\pm$ 0.0213} \\
& EEGDeformer & \underline{0.6056 {\scriptsize $\pm$ 0.0042}} & \textbf{0.6065 {\scriptsize $\pm$ 0.0052}} & \underline{0.4768 {\scriptsize $\pm$ 0.0106}} & 0.5829 {\scriptsize $\pm$ 0.0185} & \underline{0.6676 {\scriptsize $\pm$ 0.0077}} & 0.6503 {\scriptsize $\pm$ 0.0070} \\
& Stu-S & 0.5828 {\scriptsize $\pm$ 0.0085} & 0.5855 {\scriptsize $\pm$ 0.0080} & 0.4437 {\scriptsize $\pm$ 0.0113} & 0.5806 {\scriptsize $\pm$ 0.0221} & 0.6104 {\scriptsize $\pm$ 0.0324} & 0.6050 {\scriptsize $\pm$ 0.0359} \\
& Stu-M & 0.5806 {\scriptsize $\pm$ 0.0119} & 0.5818 {\scriptsize $\pm$ 0.0148} & 0.4408 {\scriptsize $\pm$ 0.0160} & 0.5846 {\scriptsize $\pm$ 0.0364} & 0.6243 {\scriptsize $\pm$ 0.0438} & 0.6190 {\scriptsize $\pm$ 0.0431} \\ \midrule
\multirow{3}{*}{\makecell{Fine-tuned\\Teachers}} 
& CBraMod & 0.6188 {\scriptsize $\pm$ 0.0072} & 0.6205 {\scriptsize $\pm$ 0.0079} & 0.4917 {\scriptsize $\pm$ 0.0096} & 0.6166 {\scriptsize $\pm$ 0.0165} & 0.6647 {\scriptsize $\pm$ 0.0266} & 0.6599 {\scriptsize $\pm$ 0.0322} \\
& LaBraM & 0.6150 {\scriptsize $\pm$ 0.0087} & 0.6152 {\scriptsize $\pm$ 0.0088} & 0.4866 {\scriptsize $\pm$ 0.0117} & 0.5997 {\scriptsize $\pm$ 0.0163} & 0.6442 {\scriptsize $\pm$ 0.0241} & 0.6464 {\scriptsize $\pm$ 0.0269} \\
& EEG-DINO & 0.5753 {\scriptsize $\pm$ 0.0080} & 0.5757 {\scriptsize $\pm$ 0.0089} & 0.4337 {\scriptsize $\pm$ 0.0108} & 0.5969 {\scriptsize $\pm$ 0.0161} & 0.6461 {\scriptsize $\pm$ 0.0214} & 0.6455 {\scriptsize $\pm$ 0.0313} \\ \midrule
\multirow{3}{*}{\makecell{KD\\(Stu-S)}} 
& FitNets & 0.5934 {\scriptsize $\pm$ 0.0066} & 0.5942 {\scriptsize $\pm$ 0.0094} & 0.4579 {\scriptsize $\pm$ 0.0089} & 0.5848 {\scriptsize $\pm$ 0.0197} & 0.6210 {\scriptsize $\pm$ 0.0281} & 0.6148 {\scriptsize $\pm$ 0.0303} \\
& Logit-std & 0.5609 {\scriptsize $\pm$ 0.0232} & 0.5666 {\scriptsize $\pm$ 0.0238} & 0.4145 {\scriptsize $\pm$ 0.0310} & 0.5623 {\scriptsize $\pm$ 0.0152} & 0.6274 {\scriptsize $\pm$ 0.0163} & 0.6295 {\scriptsize $\pm$ 0.0194} \\
& \textbf{DLink (Ours)} & 0.5979 {\scriptsize $\pm$ 0.0043} & 0.6013 {\scriptsize $\pm$ 0.0042} & 0.4638 {\scriptsize $\pm$ 0.0057} & \underline{0.6073 {\scriptsize $\pm$ 0.0133}} & 0.6488 {\scriptsize $\pm$ 0.0234} & 0.6455 {\scriptsize $\pm$ 0.0203} \\ \midrule
\multirow{3}{*}{\makecell{KD\\(Stu-M)}} 
& FitNets & 0.5831 {\scriptsize $\pm$ 0.0048} & 0.5858 {\scriptsize $\pm$ 0.0072} & 0.4441 {\scriptsize $\pm$ 0.0065} & 0.5920 {\scriptsize $\pm$ 0.0088} & 0.6218 {\scriptsize $\pm$ 0.0132} & 0.6115 {\scriptsize $\pm$ 0.0185} \\
& Logit-std & 0.5640 {\scriptsize $\pm$ 0.0122} & 0.5652 {\scriptsize $\pm$ 0.0136} & 0.4186 {\scriptsize $\pm$ 0.0162} & 0.6051 {\scriptsize $\pm$ 0.0132} & 0.6573 {\scriptsize $\pm$ 0.0154} & \underline{0.6566 {\scriptsize $\pm$ 0.0170}} \\
& \textbf{DLink (Ours)} & \textbf{0.6060 {\scriptsize $\pm$ 0.0049}} & \underline{0.6037 {\scriptsize $\pm$ 0.0052}} & 0.4666 {\scriptsize $\pm$ 0.0076} & \textbf{0.6150 {\scriptsize $\pm$ 0.0099}} & \textbf{0.6782 {\scriptsize $\pm$ 0.0071}} & \textbf{0.6676 {\scriptsize $\pm$ 0.0137}} \\ \bottomrule
\end{tabular}
}
\end{table*}

\subsection{Main Results}
\label{subsec:results}

Table~\ref{tab:main_results} summarizes the performance across four EEG benchmarks. Compared with the matched supervised students, DLink improves ACC-B for both Stu-S and Stu-M on all datasets, with broadly consistent gains on secondary metrics. Among lightweight models, DLink-M achieves the best ACC-B on all four datasets, and DLink-S remains competitive under ACC-B despite its smaller capacity. Against direct distillation baselines, DLink-M improves over FitNets and Logit-std across the reported metrics, while DLink-S remains broadly favorable with some metric-dependent variation. Overall, the main table supports DLink as an effective compact distillation strategy without implying uniform dominance on every secondary metric.

DLink also remains effective across three teacher EFMs. On FACED with Stu-S, CBraMod, LaBraM, and EEG-DINO teachers obtain ACC-B scores of $43.30\%$, $42.48\%$, and $42.58\%$, respectively, all above the supervised Stu-S baseline of $41.18\%$. Jointly tuning the CBraMod teacher on the same setting reaches $42.44\%$, below the frozen-teacher result of $43.30\%$ while adding training cost; thus, we keep teachers frozen. Detailed teacher-robustness and teacher-tuning results are provided in Appendix~\ref{app:teacher_robustness} and Appendix~\ref{app:teacher_tuning}.

\subsection{Dominant Knowledge Aggregation}

\subsubsection{Ablation of Layer Routing}

Table~\ref{tab:routing_ablation} reports routing-related ablations on FACED. \textbf{Fixed Last} injects only the final teacher representation, whereas \textbf{Fixed Avg} serves as a layer-wise distillation baseline that uniformly injects all teacher layers into the student. DLink improves over both fixed strategies by learning input-conditioned soft weights, indicating that dominant teacher knowledge should be weighted rather than manually fixed or uniformly averaged. Removing $\mathcal{L}_{\mathrm{prior}}$ or replacing the spectral-concentration score with Mean Power or Max Amplitude reduces performance, suggesting that temporal-latent energy concentration provides a useful soft cue for routing.

\begin{table}[t]
\centering
\caption{Layer-routing and spectral-concentration prior ablations on FACED.}
\label{tab:routing_ablation}
\vspace{0.4em}
\scriptsize
\setlength{\tabcolsep}{1.2pt}
\begin{tabular*}{\columnwidth}{@{\extracolsep{\fill}}l|ccc|ccc@{}}
\toprule
\multirow{2}{*}{\textbf{Variant}} & \multicolumn{3}{c|}{\textbf{Stu-S}} & \multicolumn{3}{c}{\textbf{Stu-M}} \\
& \textbf{ACC-B} & \textbf{Kappa} & \textbf{F1-W} & \textbf{ACC-B} & \textbf{Kappa} & \textbf{F1-W} \\ \midrule
w/o $\mathcal{L}_{\mathrm{prior}}$ & 0.3968 {\scriptsize $\pm$ 0.0123} & 0.3186 {\scriptsize $\pm$ 0.0134} & 0.3948 {\scriptsize $\pm$ 0.0119} & 0.4917 {\scriptsize $\pm$ 0.0107} & 0.4257 {\scriptsize $\pm$ 0.0125} & 0.4923 {\scriptsize $\pm$ 0.0146} \\
\cmidrule(lr){1-7}
Fixed Last & 0.4007 {\scriptsize $\pm$ 0.0069} & 0.3234 {\scriptsize $\pm$ 0.0080} & 0.3995 {\scriptsize $\pm$ 0.0069} & 0.4273 {\scriptsize $\pm$ 0.0093} & 0.4285 {\scriptsize $\pm$ 0.0100} & 0.4945 {\scriptsize $\pm$ 0.0088} \\
Fixed Avg & 0.4248 {\scriptsize $\pm$ 0.0114} & 0.3502 {\scriptsize $\pm$ 0.0118} & 0.4235 {\scriptsize $\pm$ 0.0095} & 0.5171 {\scriptsize $\pm$ 0.0054} & 0.4530 {\scriptsize $\pm$ 0.0067} & 0.5163 {\scriptsize $\pm$ 0.0077} \\
\cmidrule(lr){1-7}
Mean Power & 0.4273 {\scriptsize $\pm$ 0.0058} & 0.3533 {\scriptsize $\pm$ 0.0059} & \textbf{0.4259 {\scriptsize $\pm$ 0.0042}} & 0.5167 {\scriptsize $\pm$ 0.0045} & 0.4530 {\scriptsize $\pm$ 0.0065} & 0.5166 {\scriptsize $\pm$ 0.0073} \\
Max Amp & 0.4212 {\scriptsize $\pm$ 0.0094} & 0.3436 {\scriptsize $\pm$ 0.0106} & \textbf{0.4191 {\scriptsize $\pm$ 0.0101}} & 0.5036 {\scriptsize $\pm$ 0.0071} & 0.4157 {\scriptsize $\pm$ 0.0076} & \textbf{0.4819 {\scriptsize $\pm$ 0.0069}} \\
\textbf{Spectral Conc. (Ours)} & \textbf{0.4330 {\scriptsize $\pm$ 0.0132}} & 0.3548 {\scriptsize $\pm$ 0.0112} & 0.4251 {\scriptsize $\pm$ 0.0107} & \textbf{0.5220 {\scriptsize $\pm$ 0.0051}} & 0.4581 {\scriptsize $\pm$ 0.0066} & \textbf{0.5202 {\scriptsize $\pm$ 0.0073}} \\ \bottomrule
\end{tabular*}
\end{table}

\subsubsection{Layer-wise Probing and Routing Behavior}

\begin{figure*}[!t]
\centering
\begin{minipage}[t]{0.47\textwidth}
\vspace{0pt}
\centering
\includegraphics[width=\linewidth]{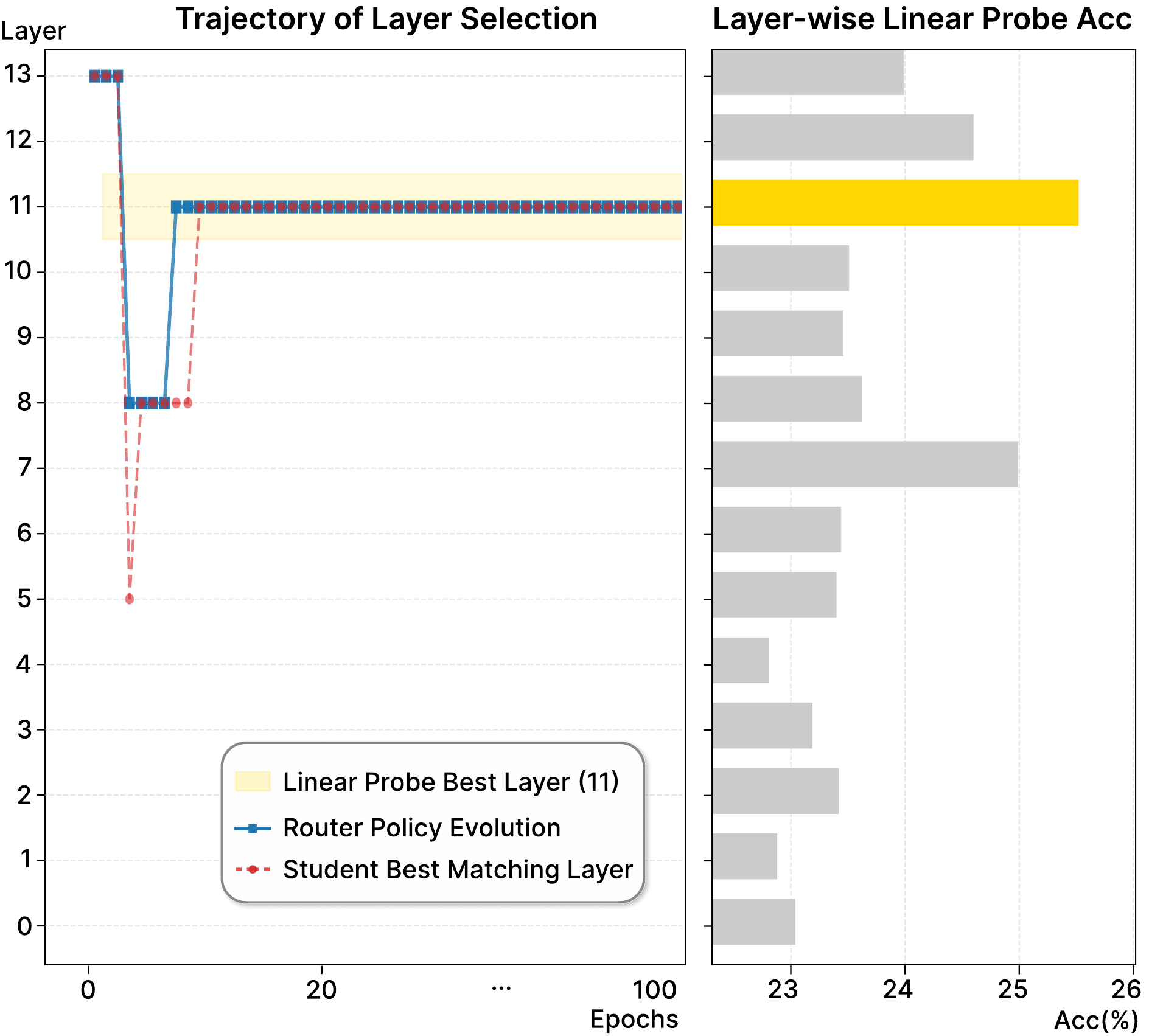}
\vspace{-1.3em}
\caption{Layer routing behavior.}
\label{fig:layer_routing}
\end{minipage}\hfill
\begin{minipage}[t]{0.49\textwidth}
\vspace{0pt}
\centering
\includegraphics[width=0.92\linewidth]{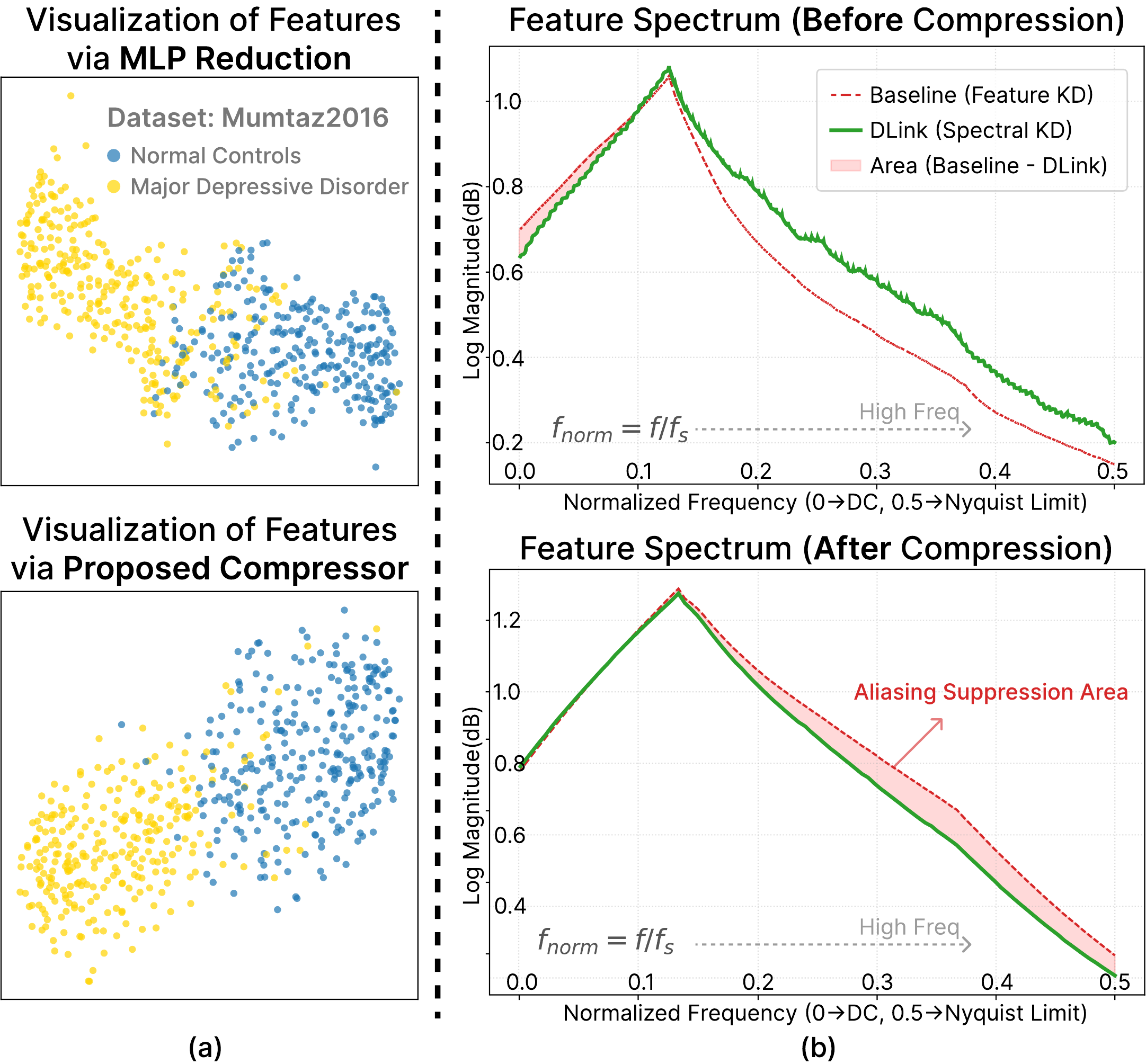}
\caption{Feature and spectrum analysis.
}
\label{fig:feature_spectrum_analysis}
\end{minipage}
\end{figure*}

At the macro level, Fig.~\ref{fig:layer_routing} compares FACED layer-wise probing with the router's policy evolution and the student's best-matching teacher layer during training. The router moves from early exploration across several layers to a stable intermediate-layer region that matches the student's closest teacher representation. This indicates that the final layer is not necessarily the most useful supervision source, and DLink can aggregate knowledge from a more informative intermediate-layer region without manually fixing the layer.

\noindent\begin{minipage}[t]{0.55\textwidth}
\vspace{0pt}
We also summarize intra-batch routing statistics using $V_b=\frac{1}{L}\sum_l \mathrm{Var}_{i\in b}(w_{i,l})$. Table~\ref{tab:residual_routing_variation} shows that routing weights retain non-zero variance across training stages. The average number of distinct top-1 layers decreases from $5.01$ in the early stage to about $2.08$ in the late stage, consistent with the epoch-level convergence toward a dominant intermediate layer reported in Appendix~\ref{app:routing_dynamics}.
\end{minipage}\hfill
\begin{minipage}[t]{0.41\textwidth}
\vspace{0pt}
\centering
\captionof{table}{Intra-batch routing statistics. \#Top-1: distinct top-1 layers per batch; Multi: batches with multiple top-1 layers.}
\label{tab:residual_routing_variation}
\vspace{0.4em}
\scriptsize
\setlength{\tabcolsep}{1.6pt}
\begin{tabular}{lcccc}
\toprule
\textbf{Stage} & $V_b$ & $\mathrm{std}(w_{11})$ & \textbf{\#Top-1} & \textbf{Multi} \\ \midrule
Early & 0.000681 & 0.0123 & 5.01 & 98.29\% \\
Middle & 0.001318 & 0.0869 & 2.06 & 82.48\% \\
Late & 0.001327 & 0.0891 & 2.08 & 82.76\% \\ \bottomrule
\end{tabular}
\end{minipage}

\subsection{Spectral Distillation Analysis}

FitNets and Logit-std in Table~\ref{tab:main_results} serve as direct feature- and logit-distillation baselines. On FACED, DLink improves ACC-B over Feature KD by $+0.91$/$+3.63$ points for Stu-S/Stu-M and over Logit KD by $+2.05$/$+0.99$ points, indicating that gains come from how teacher knowledge is aggregated and transferred; a focused comparison is provided in Appendix~\ref{app:distillation_objective}.

Fig.~\ref{fig:feature_spectrum_analysis} further visualizes the learned features. The t-SNE results show that the compressed DLink features preserve clear class separation, comparable to the full-dimensional feature before compression, indicating that compact transfer maintains discriminative structure.
The spectrum plots compare temporal-latent representation spectra before and after compression. After compression, DLink has a smaller high-frequency tail than Feature KD, indicating smoother representation spectra and less compression-induced spectral distortion, consistent with the preserved class separation.

\subsection{Compact Internalization and Inference Cost}

The router adds little training overhead because it operates on compact feature and representation-spectrum cues rather than the original high-dimensional feature. In CUDA profiling, its input shape is $(64,38,5)$, with $0.633$ ms forward latency per training batch, $0.04$M parameters, and $0.22$M FLOPs (Appendix~\ref{app:router_profile}). CPU-only batch-1 profiling on Mumtaz2016 further shows that the compact student runs in $3.653$ ms, compared with $19.345$ ms for the fine-tuned teacher, giving a $5.30\times$ speedup (Appendix~\ref{app:cpu_profile}).

\noindent\begin{minipage}[t]{0.48\textwidth}
\vspace{0pt}
Fig.~\ref{fig:efficiency_tradeoff} compares accuracy, FLOPs, and parameter volume on FACED. Under ACC-B, DLink-S is competitive with strong standalone lightweight EEG models and DLink-M achieves the best lightweight-model accuracy, while both require substantially fewer parameters and FLOPs than fully fine-tuned EFMs. The resulting models improve the accuracy--efficiency trade-off without carrying any teacher-side modules into deployment.
\end{minipage}\hfill
\begin{minipage}[t]{0.48\textwidth}
\vspace{0pt}
\centering
\includegraphics[width=\linewidth]{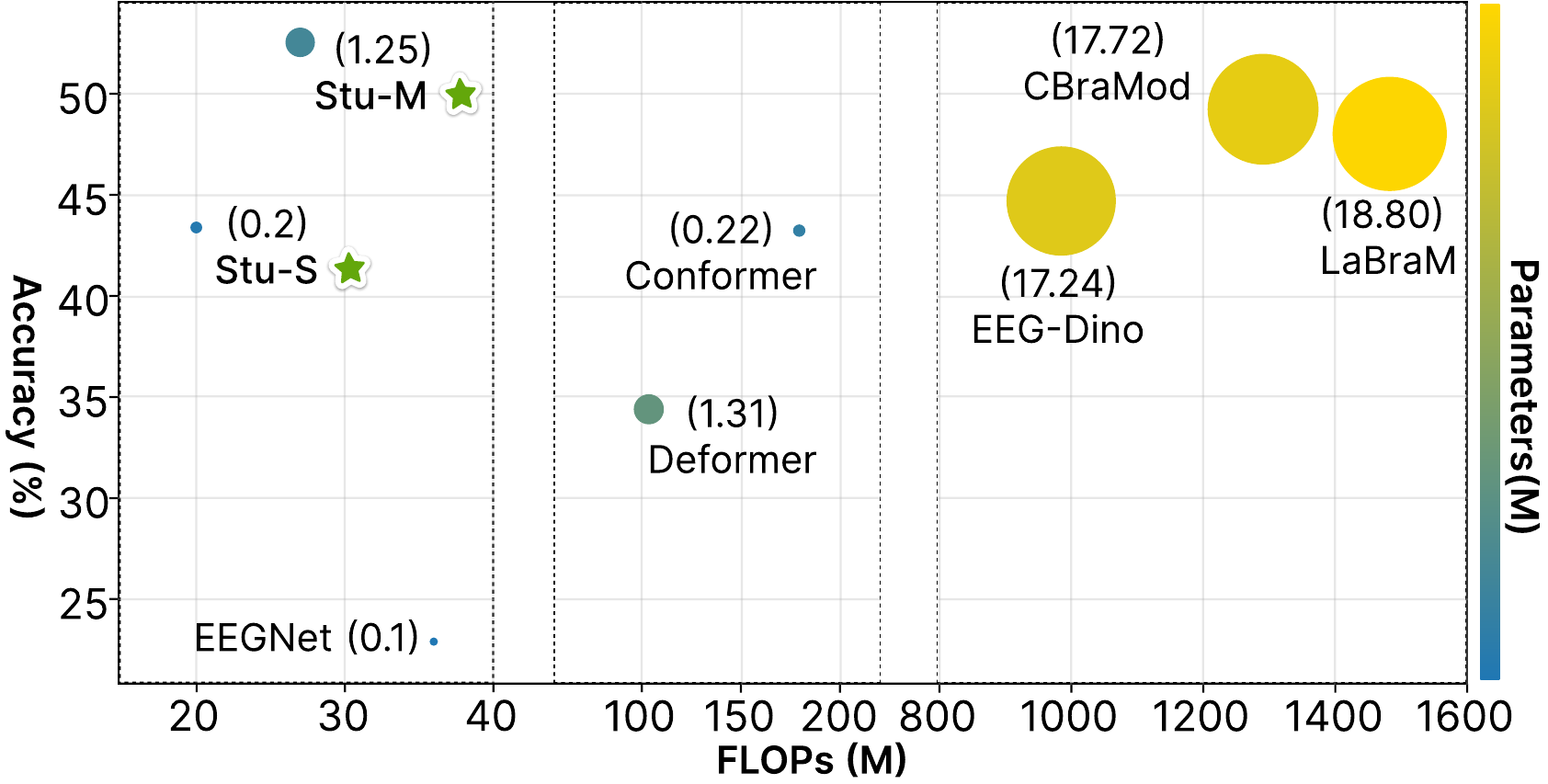}
\vspace{-1.3em}
\captionof{figure}{Accuracy--efficiency trade-off.}
\label{fig:efficiency_tradeoff}
\end{minipage}

\section{Related Work}
\label{sec:related_work}

\subsection{EEG Foundation Models}

Recent EEG foundation models (EFMs) have advanced neural representation learning by pretraining on large-scale and heterogeneous EEG corpora~\cite{codebrain,REVE}. Representative models such as EEGPT~\cite{EEGPT}, LaBraM~\cite{LarBraM}, CBraMod~\cite{CBraMod}, and EEG-DINO~\cite{EEG_DINO} learn transferable representations that can be adapted to downstream decoding tasks, and benchmarks such as AdaBrainBench~\cite{AdaBrain} further highlight their cross-subject and cross-task potential. However, EFMs usually rely on high-capacity backbones and task-specific adaptation, making it costly to retain them directly for inference. DLink therefore treats EFMs as training-time teachers and focuses on how their layer-wise knowledge can be transferred into a compact inference model.

\subsection{Knowledge Distillation and Spectral Motivation}

Knowledge distillation transfers information from a high-capacity teacher to a compact student through logit-level or feature-level supervision~\cite{SFTN,KDsurvey,DistiLLM}. Classical feature distillation encourages students to mimic teacher representations~\cite{FitNets,miniLLM}, while logit distillation matches the output distribution~\cite{Logit_std}; related ideas have also been studied in physiological and EEG settings~\cite{EmotionKD,CMCRD}. For hierarchical EFMs, however, useful decoding information may not be concentrated in the final representation. Prior layer-wise transfer and probing studies show that different pretrained layers can encode different task-relevant information~\cite{Less_is_more,combo}. DLink follows this motivation but avoids manually fixing a teacher layer by using a lightweight router to automatically aggregate dominant intermediate layers.

Frequency-domain modeling is central to EEG analysis, and recent EEG representation learning methods also exploit spectral or spectral-spatial-temporal structures~\cite{MV-SSTMA,TSception,EmT,EEGPT}. Frequency-domain objectives have further been used for representation transfer and distillation beyond EEG~\cite{FreeKD}. In convolutional networks, anti-aliased designs show that frequency-aware regularization can be useful on ordered learned feature maps before resolution reduction~\cite{Shift_Invariant,CNN_aliasing}. DLink follows this representation-level motivation by aligning magnitude and phase spectra along the temporal-latent axis as a regularizer for compact transfer. This objective is applied to learned features rather than raw EEG reconstruction, so it is used to mitigate representation-spectrum distortion during teacher--student transfer rather than to claim recovery of physiological rhythms.

\section{Conclusion}
\label{sec:conclusion}

This paper presents DLink, a compact EFM distillation framework that transfers dominant intermediate teacher knowledge through input-conditioned layer routing and representation-spectrum consistency. A project-then-compress student internalizes the routed knowledge, and both the teacher and router are discarded after training. Experiments on four EEG benchmarks show that DLink improves matched compact students under ACC-B and remains competitive with lightweight EEG baselines, providing a practical route for distilling hierarchical EFM representations into efficient inference models.
\bibliographystyle{unsrtnat}
\bibliography{Bib_DLink}

\clearpage
\input{appendix.tex}

\end{document}

%% file: appendix.tex
\appendix
\setcounter{table}{0}
\setcounter{figure}{0}
\setcounter{algorithm}{0}
\makeatletter
\@addtoreset{table}{section}
\@addtoreset{figure}{section}
\@addtoreset{algorithm}{section}
\makeatother
\renewcommand{\thetable}{\Alph{section}\arabic{table}}
\renewcommand{\thefigure}{\Alph{section}\arabic{figure}}
\renewcommand{\thealgorithm}{\Alph{section}\arabic{algorithm}}
\renewcommand{\theHtable}{appendix.table.\arabic{section}.\arabic{table}}
\renewcommand{\theHfigure}{appendix.figure.\arabic{section}.\arabic{figure}}
\providecommand{\theHalgorithm}{\arabic{algorithm}}
\renewcommand{\theHalgorithm}{appendix.algorithm.\arabic{section}.\arabic{algorithm}}
\newcommand{\AppTableCaptionSep}{\vspace{0.35em}}
\newcommand{\AppTableSetup}{\footnotesize\setlength{\tabcolsep}{3pt}\renewcommand{\arraystretch}{1.08}}
\newcommand{\AppTightTableSetup}{\footnotesize\setlength{\tabcolsep}{2pt}\renewcommand{\arraystretch}{1.08}}
\newcommand{\AppDatasetTableSetup}{\scriptsize\setlength{\tabcolsep}{1.5pt}\renewcommand{\arraystretch}{1.08}}

\section*{Technical Appendix}

\section{Evaluation Metrics}
\label{app:evaluation_metrics}

We report task metrics according to the label structure of each benchmark. For all datasets, balanced accuracy (ACC-B) is used to reduce the effect of class imbalance. For multi-class benchmarks, we additionally report weighted F1 (F1-W) and Cohen's Kappa. For binary benchmarks, we additionally report AUROC and AUC-PR.

Let $n_c$ be the number of samples in class $c$, and let $\mathrm{TP}_c$ be the number of correctly predicted samples in class $c$. Balanced accuracy is
\begin{equation}
    \mathrm{ACC\mbox{-}B}
    =
    \frac{1}{|\mathcal{Y}|}
    \sum_{c \in \mathcal{Y}}
    \frac{\mathrm{TP}_c}{n_c}.
\end{equation}
Weighted F1 averages class-wise F1 scores using class support as weights. Cohen's Kappa measures agreement between prediction and label after correcting for chance agreement. AUROC is the area under the receiver operating characteristic curve, and AUC-PR is the area under the precision--recall curve.

\begin{center}
\begin{minipage}{\textwidth}
\centering
\captionof{table}{Evaluation metrics reported for each benchmark according to task label structure.}
\label{tab:appendix_metrics}
\AppTableCaptionSep
{\AppTableSetup
\begin{tabular*}{\textwidth}{@{\extracolsep{\fill}}llll@{}}
\toprule
\textbf{Dataset} & \textbf{Task} & \textbf{Main metric} & \textbf{Additional metrics} \\ \midrule
FACED & 9-class emotion recognition & ACC-B & F1-W, Kappa \\
Mumtaz2016 & Binary depression diagnosis & ACC-B & AUROC, AUC-PR \\
PhysioNet-MI & 4-class motor imagery & ACC-B & F1-W, Kappa \\
SHU-MI & Binary motor imagery & ACC-B & AUROC, AUC-PR \\ \bottomrule
\end{tabular*}}
\end{minipage}
\end{center}

\section{Dataset Details and Preprocessing}
\label{app:dataset_details}

Table~\ref{tab:dataset_preprocessing} summarizes the downstream benchmark statistics, and Table~\ref{tab:split_leakage_control} reports the corresponding train/validation/test partitions. We follow the CBraMod downstream setting~\cite{CBraMod}: EEG signals are resampled to 200 Hz, represented as 1-second patches with 200 time points per patch, and then grouped into task-specific examples. Splitting is performed before patching or windowing, so all patches/windows derived from the same original subject, session, recording, or trial container remain in a single split.

\begin{center}
\begin{minipage}{\textwidth}
\centering
\captionof{table}{Dataset statistics for the downstream benchmarks. Original rate, duration, channel number, and sample count follow the CBraMod downstream benchmark table.}
\label{tab:dataset_preprocessing}
\AppTableCaptionSep
{\AppDatasetTableSetup
\begin{tabular}{@{}>{\raggedright\arraybackslash}p{0.135\textwidth}
>{\centering\arraybackslash}p{0.080\textwidth}
>{\raggedright\arraybackslash}p{0.245\textwidth}
>{\centering\arraybackslash}p{0.120\textwidth}
>{\centering\arraybackslash}p{0.075\textwidth}
>{\centering\arraybackslash}p{0.100\textwidth}
>{\centering\arraybackslash}p{0.115\textwidth}@{}}
\toprule
\textbf{Dataset} & \makecell[c]{\textbf{\#}\\\textbf{Subj.}} & \textbf{Trials/Sessions} & \makecell[c]{\textbf{Original}\\\textbf{rate}} & \makecell[c]{\textbf{\#}\\\textbf{Ch.}} & \textbf{Duration} & \makecell[c]{\textbf{\#}\\\textbf{Samples}} \\ \midrule
FACED & 123 & 28 video trials/subject & 250 Hz & 32 & 10 s & 10,332 \\
Mumtaz2016 & 64 & 128 resting EDF recordings; task files excluded & 256 Hz & 19 & 5 s & 7,143 \\
PhysioNet-MI & 109 & 6 selected MI runs/subject & 160 Hz & 64 & 4 s & 9,837 \\
SHU-MI & 25 & 5 sessions/subject & 250 Hz & 32 & 4 s & 11,988 \\ \bottomrule
\end{tabular}
}
\end{minipage}
\end{center}

\begin{center}
\begin{minipage}{\textwidth}
\centering
\captionof{table}{Train/validation/test partitions for each downstream benchmark.}
\label{tab:split_leakage_control}
\AppTableCaptionSep
{\AppTableSetup
\begin{tabular}{@{}>{\raggedright\arraybackslash}p{0.22\textwidth}
>{\raggedright\arraybackslash}p{0.68\textwidth}@{}}
\toprule
\textbf{Dataset} & \textbf{Train/Val/Test split} \\ \midrule
FACED & 80/20/23 subjects \\
Mumtaz2016 & Class-stratified resting-recording split following CBraMod: healthy files 40/8/remaining; MDD files 42/10/remaining \\
PhysioNet-MI & 70/19/20 subjects \\
SHU-MI & 15/5/5 subjects (75/25/25 session files) \\ \bottomrule
\end{tabular}
}
\end{minipage}
\end{center}

\subsection{Existing Assets and Licenses}
\label{app:asset_licenses}

We use existing datasets, teacher EFMs, and baseline methods only under their original access terms and cite the corresponding papers in the main text. We do not redistribute restricted raw EEG data. Users of any accompanying DLink code should obtain each dataset from its official source and follow the corresponding license, data-use agreement, and citation requirements.

\begin{center}
\begin{minipage}{\textwidth}
\centering
\captionof{table}{Dataset sources and license or usage terms.}
\label{tab:asset_licenses}
\AppTableCaptionSep
{\AppDatasetTableSetup
\begin{tabular}{@{}>{\raggedright\arraybackslash}p{0.15\textwidth}
>{\raggedright\arraybackslash}p{0.49\textwidth}
>{\raggedright\arraybackslash}p{0.28\textwidth}@{}}
\toprule
\textbf{Dataset} & \textbf{Source} & \textbf{License / usage terms} \\ \midrule
FACED & FACED dataset, Synapse project \texttt{syn50614194}; \url{https://www.nature.com/articles/s41597-023-02650-w} & CC-BY metadata; the Scientific Data article is CC BY 4.0, with download terms governed by the Synapse page. \\
SHU-MI & figshare ``\texttt{shu\_dataset}'', DOI 10.6084/m9.figshare.19228725.v1; \url{https://doi.org/10.6084/m9.figshare.19228725.v1} & CC BY 4.0. \\
PhysioNet-MI & PhysioNet EEG Motor Movement/Imagery Dataset; \url{https://www.physionet.org/content/eegmmidb/} & Open Data Commons Attribution License v1.0. \\
Mumtaz2016 & figshare MDD Patients and Healthy Controls EEG Data, DOI 10.6084/m9.figshare.4244171; \url{https://figshare.com/articles/dataset/EEG_Data_New/4244171} & CC BY 4.0. \\ \bottomrule
\end{tabular}}
\end{minipage}
\end{center}

\section{Architecture and Algorithm Details}
\label{app:architecture_algorithm}

This section provides the implementation-level details needed to reproduce the compact student, the router, and the training/inference procedures. The emphasis is on the components that determine the distillation pathway, not on introducing an additional architectural contribution.

\subsection{Compact Project-then-Compress Student}
\label{app:student_structure}

The compact student is an implementation vehicle for internalizing routed teacher knowledge, not a standalone architectural contribution. Table~\ref{tab:student_architecture} summarizes its main stages in a layout-agnostic way. Since different EFMs may expose different tokenization schemes or latent layouts, the student projection is matched to the teacher representation used for distillation rather than assuming a fixed EEG tensor shape.

\begin{table}[!htbp]
\centering
\caption{Project-then-compress student architecture used for DLink. The table reports the main implementation stages rather than an exhaustive layer-by-layer specification.}
\label{tab:student_architecture}
\AppTableCaptionSep
\AppTableSetup
\resizebox{\textwidth}{!}{
\begin{tabular}{llll}
\toprule
\textbf{Stage} & \textbf{Operation} & \textbf{Configuration} & \textbf{Notes} \\ \midrule
Projection & Depthwise separable temporal CNN & kernel $35$ & Local temporal feature extraction \\
Projection & Transformer encoder over latent units & model dimension matched to feature dimension & Long-range context modeling \\
Projection & Learnable CNN/Transformer fusion & scalar gate $\alpha$ & Produces feature $f_S^p$ for distillation \\
Compression & Latent-resolution compression & computed kernel/stride when applicable & Stronger compression for Stu-S \\
Compression & Spatio-temporal CNN downsampling & \makecell[l]{Stu-S: $15/15$, $C_o=15$\\Stu-M: $20/10$, $C_o=30$} & Structured compact transfer \\
Prediction & MLP classifier & \makecell[l]{Stu-S: hidden 100\\Stu-M: hidden 200} & Compact task prediction \\ \bottomrule
\end{tabular}}
\end{table}

The two student scales differ mainly in compression strength and hidden dimension. Stu-S uses one Transformer block, hidden dimension 100, temporal kernel/stride $15/15$, output channels 15, and stronger latent-resolution compression. Stu-M uses three Transformer blocks, hidden dimension 200, temporal kernel/stride $20/10$, output channels 30, and keeps the latent resolution by default.

We ablate the CNN and Transformer branches to check whether both projection components contribute to compact transfer. Table~\ref{tab:student_structure_ablation} reports the FACED results under the same student scales used in the main experiments.

\begin{table}[!htbp]
\centering
\caption{Ablation of the student projection components on FACED. Values are mean $\pm$ standard deviation over five seeds.}
\label{tab:student_structure_ablation}
\AppTableCaptionSep
\AppTightTableSetup
\resizebox{\textwidth}{!}{
\begin{tabular}{l|ccc|ccc}
\toprule
\multirow{2}{*}{\textbf{Variant}} & \multicolumn{3}{c|}{\textbf{Stu-S}} & \multicolumn{3}{c}{\textbf{Stu-M}} \\
& \textbf{ACC-B} & \textbf{Kappa} & \textbf{F1-W} & \textbf{ACC-B} & \textbf{Kappa} & \textbf{F1-W} \\ \midrule
w/o CNN & 0.3933 {\scriptsize $\pm$ 0.0049} & 0.3134 {\scriptsize $\pm$ 0.0053} & 0.3897 {\scriptsize $\pm$ 0.0042} & 0.5086 {\scriptsize $\pm$ 0.0068} & 0.4447 {\scriptsize $\pm$ 0.0061} & 0.5096 {\scriptsize $\pm$ 0.0034} \\
w/o Trans & 0.4224 {\scriptsize $\pm$ 0.0226} & 0.3459 {\scriptsize $\pm$ 0.0250} & 0.4166 {\scriptsize $\pm$ 0.0232} & 0.4913 {\scriptsize $\pm$ 0.0082} & 0.4245 {\scriptsize $\pm$ 0.0098} & 0.4898 {\scriptsize $\pm$ 0.0090} \\
\textbf{DLink (Ours)} & \textbf{0.4330 {\scriptsize $\pm$ 0.0132}} & \textbf{0.3548 {\scriptsize $\pm$ 0.0112}} & \textbf{0.4251 {\scriptsize $\pm$ 0.0107}} & \textbf{0.5220 {\scriptsize $\pm$ 0.0051}} & \textbf{0.4581 {\scriptsize $\pm$ 0.0066}} & \textbf{0.5202 {\scriptsize $\pm$ 0.0073}} \\ \bottomrule
\end{tabular}}
\end{table}

The full projection block performs best for both student scales, indicating that local temporal extraction and latent-unit context modeling provide complementary support for distillation.

\subsection{Layer Router Architecture}
\label{app:router_architecture}

The router predicts input-conditioned soft weights over teacher layers. Its input is constructed from compact student-feature cues and a representation-spectrum cue, rather than from the original high-dimensional feature map. In the profiled setting, the router input shape is $(64,38,5)$.

\begin{table}[!htbp]
\centering
\caption{Layer router architecture. The router has $0.04$M parameters and $0.22$M FLOPs in our profiling.}
\label{tab:router_architecture}
\AppTableCaptionSep
\AppTableSetup
\resizebox{\columnwidth}{!}{
\begin{tabular}{llll}
\toprule
\textbf{Component} & \textbf{Operation} & \textbf{Configuration} & \textbf{Notes} \\ \midrule
Input normalization & LayerNorm & compact feature/spectrum cue & Applied before routing blocks \\
Channel fuser & Conv1D + BatchNorm + GELU & router hidden dim $64$ & Fuses channel-wise cues \\
Sequence processor & Transformer encoder + LayerNorm & 2 attention heads & Models dependencies in routing input \\
Pooling & Adaptive average pooling & global pooling & Produces global routing summary \\
Routing head & Linear projection & output dimension $L$ & Layer-weight logits \\
Layer weights & Temperature softmax & temperature $\tau$ & $\mathrm{softmax}(a/\tau)$ \\ \bottomrule
\end{tabular}}
\end{table}

The router is lightweight because it processes compact cues rather than the teacher's full feature map. It is therefore suitable as a training-time module that can be discarded after the student has internalized the routed supervision.

\subsection{DLink Training Algorithm}
\label{app:training_algorithm}

Algorithm~\ref{alg:dlink_training} summarizes one DLink training step. The teacher provides frozen layer-wise representations, while the student and router are optimized jointly.

\begin{algorithm}[H]
\caption{DLink training procedure}
\label{alg:dlink_training}
\begin{algorithmic}[1]
\REQUIRE Frozen teacher EFM $\mathcal{T}$, compact student $\mathcal{S}$, layer router $R$, training set $\mathcal{D}$
\FOR{each mini-batch $(X,y)$}
    \STATE Extract the routed teacher hidden-state list $\{f_T^{(l)}\}_{l=1}^{L}$ using frozen $\mathcal{T}$
    \STATE Compute student feature $f_S^{p}$, compressed feature $f_S^{c}$, and prediction $\hat{y}_S$
    \STATE Construct router input $z_{\mathrm{in}}$ from pooled student features and representation-spectrum cues
    \STATE Predict layer weights $w=\mathrm{softmax}(R(z_{\mathrm{in}})/\tau)$
    \STATE Aggregate routed teacher knowledge $f_T^{*}=\sum_{l=1}^{L}w_l f_T^{(l)}$
    \STATE Compute classification loss $\mathcal{L}_{\mathrm{cls}}$
    \STATE Compute spectral distillation loss $\mathcal{L}_{\mathrm{kd}}$
    \STATE Compute spectral-concentration routing prior loss $\mathcal{L}_{\mathrm{prior}}$
    \STATE Update $\mathcal{S}$ and $R$ with $\mathcal{L}_{\mathrm{cls}}+\lambda\mathcal{L}_{\mathrm{kd}}+\mathcal{L}_{\mathrm{prior}}$
\ENDFOR
\STATE Discard $\mathcal{T}$ and $R$ after training
\RETURN Compact student $\mathcal{S}$
\end{algorithmic}
\end{algorithm}

After training, both the teacher and the router are removed. The returned model is therefore the compact student alone.

\subsection{DLink Inference Algorithm}
\label{app:inference_algorithm}

Algorithm~\ref{alg:dlink_inference} shows the resulting inference path. No teacher forward pass, routing computation, or distillation loss is used during deployment.

\begin{algorithm}[H]
\caption{DLink inference procedure}
\label{alg:dlink_inference}
\begin{algorithmic}[1]
\REQUIRE Trained compact student $\mathcal{S}$, EEG sample $X$
\STATE Compute prediction $\hat{y}=\mathcal{S}(X)$
\RETURN $\hat{y}$
\end{algorithmic}
\end{algorithm}

Thus, the inference cost is identical to a standalone compact student with the same architecture.

\section{Implementation Details}
\label{app:implementation_details}

All models are implemented in PyTorch and optimized with AdamW. Table~\ref{tab:training_hyperparameters} reports the training hyperparameters used in the main experiments. For each run, we select the checkpoint with the highest validation accuracy and evaluate it on the test set.

Each DLink run uses one teacher EFM at a time. The teacher encoder is kept frozen while training the compact student and router. DLink uses the routed hidden-state list exposed by the frozen teacher encoder and does not require a teacher-side classifier. Different teacher EFMs are evaluated in separate runs.

\begin{table}[!htbp]
\centering
\caption{Training hyperparameters used in the main DLink experiments. AdamW uses $\beta_1=0.9$ and $\beta_2=0.999$.}
\label{tab:training_hyperparameters}
\AppTableCaptionSep
\AppTableSetup
\resizebox{\textwidth}{!}{
\begin{tabular}{llllllll}
\toprule
\textbf{Dataset} & \textbf{LR} & \textbf{Batch} & \textbf{Epochs} & \textbf{Optimizer} & \textbf{Weight decay} & \textbf{Router $\tau$} & \textbf{Seeds} \\ \midrule
FACED & $2\times10^{-3}$ & 64 & 100 & AdamW & 0.05 & 2.0 & 5 \\
Mumtaz2016 & $8\times10^{-3}$ & 64 & 100 & AdamW & 0.05 & 2.0 & 5 \\
PhysioNet-MI & $2\times10^{-3}$ & 64 & 100 & AdamW & 0.05 & 2.0 & 5 \\
SHU-MI & $5\times10^{-4}$ & 64 & 100 & AdamW & 0.05 & 2.0 & 5 \\ \bottomrule
\end{tabular}}
\end{table}

The learning rate and spectral distillation weight $\lambda$ are selected by validation-set grid search. The router temperature $\tau$ follows the standard temperature-scaled softmax formulation used in knowledge distillation~\cite{HintonKD}; following this practice of using a temperature above the ordinary softmax scale, we fix $\tau=2.0$ for all datasets. We use the same value for the guidance temperature, i.e., $\tau_q=\tau$, so that the learned routing distribution and the spectral-concentration guidance distribution are compared at the same softmax scale. These temperatures are used only during training because the router is discarded after distillation.

\subsection{Compute Resources}
\label{app:compute_resources}

All experiments were conducted on a server equipped with four NVIDIA A100-SXM4-80GB GPUs, providing 320 GB of total GPU memory. The server uses dual Intel Xeon Platinum 8452Y CPUs with 72 physical cores and 144 logical threads, and has approximately 512 GB of system memory. The software environment uses Python 3.9.23 and PyTorch 2.5.1+cu121; PyTorch was built against CUDA 12.1 with cuDNN 9.1.0.

The model-scale compute cost is reported through the parameter and FLOP comparisons in Fig.~\ref{fig:efficiency_tradeoff}. The training-time router cost and CPU-only inference latency are separately profiled in Appendix~\ref{app:profiling_protocol}. These measurements are intended to support the reported efficiency claim and do not include EEG acquisition, online preprocessing, communication, or operating-system scheduling overhead.

\section{Teacher Robustness and Hyperparameter Analyses}
\label{app:additional_teacher_analysis}
\label{app:additional_results}

This section reports experiments that vary the teacher or training hyperparameters while keeping the compact student design unchanged. These analyses clarify how teacher selection, teacher-side tuning, and validation-based hyperparameter choices affect DLink training rather than the inference architecture.

\subsection{Robustness to Different Teacher EFMs}
\label{app:teacher_robustness}

We first test whether DLink depends on a specific EFM teacher. For each run, the teacher is fixed to one EFM, and the routed hidden-state list is constructed within that teacher. Table~\ref{tab:teacher_ablation_all} shows that DLink remains effective with CBraMod, LaBraM, and EEG-DINO teachers, although the strongest teacher varies across datasets and student scales.

\begin{table}[!htbp]
\centering
\caption{Performance of DLink students distilled from three EFM teachers across four downstream benchmarks. Values are mean $\pm$ standard deviation over five seeds; the best result within each student scale and dataset is highlighted in \textbf{bold}.}
\label{tab:teacher_ablation_all}
\AppTableCaptionSep
\AppTightTableSetup
\resizebox{\textwidth}{!}{
\begin{tabular}{ll|ccc|ccc}
\toprule
\multicolumn{2}{c|}{\textbf{Configuration}} & \multicolumn{3}{c|}{\textbf{FACED (9-class)}} & \multicolumn{3}{c}{\textbf{Mumtaz2016 (2-class)}} \\ \cmidrule{1-8}
\textbf{Student} & \textbf{Teacher} & \textbf{ACC-B} & \textbf{F1-W} & \textbf{Kappa} & \textbf{ACC-B} & \textbf{AUROC} & \textbf{AUC-PR} \\ \midrule
\multirow{3}{*}{Stu-S} & CBraMod & \textbf{0.4330} {\scriptsize $\pm$ 0.0132} & \textbf{0.4251} {\scriptsize $\pm$ 0.0107} & \textbf{0.3548} {\scriptsize $\pm$ 0.0113} & 0.8970 {\scriptsize $\pm$ 0.0094} & 0.9571 {\scriptsize $\pm$ 0.0083} & 0.9655 {\scriptsize $\pm$ 0.0065} \\
 & LaBraM & 0.4248 {\scriptsize $\pm$ 0.0063} & 0.4233 {\scriptsize $\pm$ 0.0080} & 0.3498 {\scriptsize $\pm$ 0.0074} & \textbf{0.9010} {\scriptsize $\pm$ 0.0045} & 0.9612 {\scriptsize $\pm$ 0.0044} & \textbf{0.9683} {\scriptsize $\pm$ 0.0034} \\
 & EEG-DINO & 0.4258 {\scriptsize $\pm$ 0.0171} & 0.4227 {\scriptsize $\pm$ 0.0172} & 0.3500 {\scriptsize $\pm$ 0.0190} & 0.9008 {\scriptsize $\pm$ 0.0069} & \textbf{0.9664} {\scriptsize $\pm$ 0.0074} & 0.9722 {\scriptsize $\pm$ 0.0057} \\ \midrule
\multirow{3}{*}{Stu-M} & CBraMod & 0.5221 {\scriptsize $\pm$ 0.0052} & 0.5202 {\scriptsize $\pm$ 0.0074} & 0.4581 {\scriptsize $\pm$ 0.0066} & 0.9015 {\scriptsize $\pm$ 0.0041} & 0.9671 {\scriptsize $\pm$ 0.0082} & 0.9727 {\scriptsize $\pm$ 0.0063} \\
 & LaBraM & 0.5191 {\scriptsize $\pm$ 0.0096} & 0.5194 {\scriptsize $\pm$ 0.0110} & 0.4563 {\scriptsize $\pm$ 0.0109} & \textbf{0.9024} {\scriptsize $\pm$ 0.0082} & \textbf{0.9695} {\scriptsize $\pm$ 0.0074} & \textbf{0.9738} {\scriptsize $\pm$ 0.0060} \\
 & EEG-DINO & \textbf{0.5246} {\scriptsize $\pm$ 0.0074} & \textbf{0.5236} {\scriptsize $\pm$ 0.0075} & \textbf{0.4616} {\scriptsize $\pm$ 0.0085} & 0.8999 {\scriptsize $\pm$ 0.0143} & 0.9628 {\scriptsize $\pm$ 0.0115} & 0.9686 {\scriptsize $\pm$ 0.0096} \\ \midrule \midrule
\multicolumn{2}{c|}{\textbf{Configuration}} & \multicolumn{3}{c|}{\textbf{PhysioNet-MI (4-class)}} & \multicolumn{3}{c}{\textbf{SHU-MI (2-class)}} \\ \cmidrule{1-8}
\textbf{Student} & \textbf{Teacher} & \textbf{ACC-B} & \textbf{F1-W} & \textbf{Kappa} & \textbf{ACC-B} & \textbf{AUROC} & \textbf{AUC-PR} \\ \midrule
\multirow{3}{*}{Stu-S} & CBraMod & 0.5915 {\scriptsize $\pm$ 0.0009} & 0.5950 {\scriptsize $\pm$ 0.0028} & 0.4554 {\scriptsize $\pm$ 0.0013} & \textbf{0.6073} {\scriptsize $\pm$ 0.0133} & \textbf{0.6488} {\scriptsize $\pm$ 0.0234} & 0.6455 {\scriptsize $\pm$ 0.0203} \\
 & LaBraM & \textbf{0.5979} {\scriptsize $\pm$ 0.0043} & \textbf{0.6013} {\scriptsize $\pm$ 0.0042} & \textbf{0.4638} {\scriptsize $\pm$ 0.0057} & 0.6054 {\scriptsize $\pm$ 0.0150} & 0.6487 {\scriptsize $\pm$ 0.0165} & \textbf{0.6492} {\scriptsize $\pm$ 0.0193} \\
 & EEG-DINO & 0.5937 {\scriptsize $\pm$ 0.0080} & 0.5961 {\scriptsize $\pm$ 0.0089} & 0.4582 {\scriptsize $\pm$ 0.0107} & 0.5924 {\scriptsize $\pm$ 0.0165} & 0.6303 {\scriptsize $\pm$ 0.0171} & 0.6289 {\scriptsize $\pm$ 0.0231} \\ \midrule
\multirow{3}{*}{Stu-M} & CBraMod & 0.5954 {\scriptsize $\pm$ 0.0094} & 0.5970 {\scriptsize $\pm$ 0.0102} & 0.4605 {\scriptsize $\pm$ 0.0126} & \textbf{0.6150} {\scriptsize $\pm$ 0.0099} & \textbf{0.6782} {\scriptsize $\pm$ 0.0071} & \textbf{0.6676} {\scriptsize $\pm$ 0.0137} \\
 & LaBraM & \textbf{0.6060} {\scriptsize $\pm$ 0.0049} & \textbf{0.6037} {\scriptsize $\pm$ 0.0052} & \textbf{0.4666} {\scriptsize $\pm$ 0.0076} & 0.6068 {\scriptsize $\pm$ 0.0155} & 0.6571 {\scriptsize $\pm$ 0.0267} & 0.6506 {\scriptsize $\pm$ 0.0299} \\
 & EEG-DINO & 0.5977 {\scriptsize $\pm$ 0.0041} & 0.5990 {\scriptsize $\pm$ 0.0046} & 0.4635 {\scriptsize $\pm$ 0.0055} & 0.5988 {\scriptsize $\pm$ 0.0172} & 0.6402 {\scriptsize $\pm$ 0.0295} & 0.6370 {\scriptsize $\pm$ 0.0261} \\ \bottomrule
\end{tabular}}
\end{table}

These results support the use of a frozen EFM as a training-time representation source rather than a teacher-specific inference component. They also show why teacher-side design choices should be reported explicitly, because the best teacher is not identical across all benchmarks.

\subsection{Frozen vs. Fine-tuned Teacher}
\label{app:teacher_tuning}

We additionally evaluate whether unfreezing the teacher improves distillation. This experiment uses Stu-S with CBraMod as the teacher. The student-side optimization follows the main experiments, while the teacher-side parameters are unfrozen and trained with a learning rate set to $0.001$ times the main learning rate. Overall, teacher-side fine-tuning does not yield clear or consistent gains over the frozen-teacher setting while increasing training cost.

\begin{table}[!htbp]
\centering
\caption{Effect of teacher-side fine-tuning with Stu-S and CBraMod across four datasets. Values are mean $\pm$ standard deviation over five seeds; the frozen-teacher row uses the corresponding CBraMod results from Table~\ref{tab:teacher_ablation_all}.}
\label{tab:teacher_tuning}
\AppTableCaptionSep
\AppTightTableSetup
\resizebox{\textwidth}{!}{
\begin{tabular}{l|ccc|ccc}
\toprule
\multirow{2}{*}{\textbf{Setting}} & \multicolumn{3}{c|}{\textbf{FACED (9-class)}} & \multicolumn{3}{c}{\textbf{Mumtaz2016 (2-class)}} \\
& \textbf{ACC-B} & \textbf{F1-W} & \textbf{Kappa} & \textbf{ACC-B} & \textbf{AUROC} & \textbf{AUC-PR} \\ \midrule
CBraMod fine-tuned & 0.4244 {\scriptsize $\pm$ 0.0088} & 0.4220 {\scriptsize $\pm$ 0.0108} & 0.3492 {\scriptsize $\pm$ 0.0107} & 0.9023 {\scriptsize $\pm$ 0.0187} & 0.9705 {\scriptsize $\pm$ 0.0082} & 0.9752 {\scriptsize $\pm$ 0.0060} \\
CBraMod frozen & 0.4330 {\scriptsize $\pm$ 0.0132} & 0.4251 {\scriptsize $\pm$ 0.0107} & 0.3548 {\scriptsize $\pm$ 0.0113} & 0.8970 {\scriptsize $\pm$ 0.0094} & 0.9571 {\scriptsize $\pm$ 0.0083} & 0.9655 {\scriptsize $\pm$ 0.0065} \\ \midrule \midrule
\multirow{2}{*}{\textbf{Setting}} & \multicolumn{3}{c|}{\textbf{PhysioNet-MI (4-class)}} & \multicolumn{3}{c}{\textbf{SHU-MI (2-class)}} \\
& \textbf{ACC-B} & \textbf{F1-W} & \textbf{Kappa} & \textbf{ACC-B} & \textbf{AUROC} & \textbf{AUC-PR} \\ \midrule
CBraMod fine-tuned & 0.5910 {\scriptsize $\pm$ 0.0072} & 0.5931 {\scriptsize $\pm$ 0.0071} & 0.4546 {\scriptsize $\pm$ 0.0096} & 0.5761 {\scriptsize $\pm$ 0.0245} & 0.6235 {\scriptsize $\pm$ 0.0429} & 0.6256 {\scriptsize $\pm$ 0.0393} \\
CBraMod frozen & 0.5915 {\scriptsize $\pm$ 0.0009} & 0.5950 {\scriptsize $\pm$ 0.0028} & 0.4554 {\scriptsize $\pm$ 0.0013} & 0.6073 {\scriptsize $\pm$ 0.0133} & 0.6488 {\scriptsize $\pm$ 0.0234} & 0.6455 {\scriptsize $\pm$ 0.0203} \\ \bottomrule
\end{tabular}}
\end{table}

Teacher-side fine-tuning improves some binary-diagnosis metrics on Mumtaz2016, but it reduces or leaves unchanged the main balanced-accuracy result on the other datasets. We therefore keep the teacher frozen in the main protocol to avoid extra training cost without a consistent accuracy gain.

\subsection{Hyperparameter Sensitivity}
\label{app:hyperparameter_sensitivity}

\begin{figure}[!htbp]
\centering
\includegraphics[width=\textwidth]{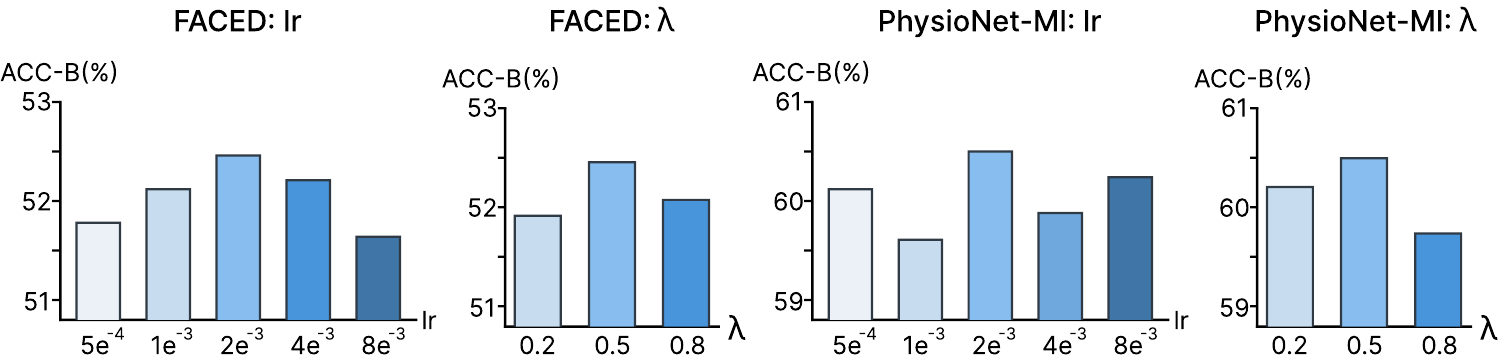}
\caption{Validation-guided sensitivity analysis for learning rate and spectral distillation weight $\lambda$ on FACED and PhysioNet-MI.}
\label{fig:hyperparameter_sensitivity}
\end{figure}

We analyze sensitivity to the learning rate and spectral distillation weight $\lambda$ on FACED and PhysioNet-MI. As shown in Fig.~\ref{fig:hyperparameter_sensitivity}, performance remains stable within a reasonable neighborhood of the selected settings.

The sensitivity curves are used to check whether the selected settings are isolated optima. The observed stability around the chosen learning rates and $\lambda$ values supports the validation-grid protocol used for the main experiments.

\section{Additional Routing Dynamics}
\label{app:routing_dynamics}

This section expands the routing analysis in the main text by reporting the numerical layer statistics behind Fig.~\ref{fig:layer_routing}. The goal is to show how the router moves from broad early exploration to a concentrated intermediate-layer policy.

\begin{table}[!htbp]
\centering
\caption{Epoch-level routing dynamics on FACED. Average-MSE is computed between student features and each teacher layer representation; lower values indicate closer matching.}
\label{tab:epoch_routing_dynamics}
\AppTableCaptionSep
\AppTightTableSetup
\resizebox{\textwidth}{!}{
\begin{tabular}{clll}
\toprule
\textbf{Epoch} & \textbf{Average router-weight top-3} & \textbf{Lowest average-MSE top-3} & \textbf{Top-1 router distribution} \\ \midrule
0 & $L_{13}=0.154$, $L_{0}=0.078$, $L_{10}=0.066$ & $L_{13}=1.554$, $L_{0}=1.761$, $L_{5}=1.812$ & $L_{13}$ 89.1\%, $L_{0}$ 4.0\%, $L_{10}$ 1.5\% \\
2 & $L_{13}=0.083$, $L_{5}=0.078$, $L_{8}=0.077$ & $L_{13}=1.540$, $L_{5}=1.562$, $L_{8}=1.573$ & $L_{13}$ 36.0\%, $L_{5}$ 17.1\%, $L_{8}$ 13.0\% \\
4 & $L_{8}=0.082$, $L_{11}=0.081$, $L_{7}=0.080$ & $L_{8}=1.485$, $L_{7}=1.488$, $L_{5}=1.491$ & $L_{8}$ 29.4\%, $L_{11}$ 24.8\%, $L_{7}$ 17.5\% \\
9 & $L_{11}=0.104$, $L_{8}=0.087$, $L_{7}=0.084$ & $L_{11}=1.350$, $L_{8}=1.383$, $L_{7}=1.387$ & $L_{11}$ 96.2\%, $L_{0}$ 2.8\% \\
49 & $L_{11}=0.595$, $L_{10}=0.039$, $L_{12}=0.036$ & $L_{11}=0.743$, $L_{10}=1.238$, $L_{12}=1.318$ & $L_{11}$ 97.7\%, $L_{0}$ 1.7\%, $L_{13}$ 0.5\% \\
99 & $L_{11}=0.605$, $L_{10}=0.038$, $L_{12}=0.035$ & $L_{11}=0.712$, $L_{10}=1.224$, $L_{12}=1.303$ & $L_{11}$ 97.8\%, $L_{0}$ 1.8\%, $L_{13}$ 0.4\% \\ \bottomrule
\end{tabular}}
\end{table}

The epoch-level statistics show that $L_{11}$ becomes both the highest-weighted layer and the closest layer by average MSE in late training. We also inspect mini-batches to ensure that this aggregate concentration does not hide all input-level variation.

\begin{table}[!htbp]
\centering
\caption{Representative examples of residual routing variation within late-epoch mini-batches on FACED.}
\label{tab:batch_routing_examples}
\AppTableCaptionSep
{\AppTableSetup
\begin{tabular*}{\columnwidth}{@{\extracolsep{\fill}}llll@{}}
\toprule
\textbf{Epoch / batch} & \textbf{Sample} & \textbf{Top-1 layer} & \textbf{Notable weights} \\ \midrule
99 / 0 & 0 & $L_{11}$ & $w_{11}=0.6008$ \\
99 / 0 & 5 & $L_{13}$ & $w_{13}=0.7702$, $w_{11}=0.0052$ \\
99 / 0 & 43 & $L_{0}$ & $w_{0}=0.3679$ \\
99 / 0 & 62 & $L_{7}$ & $w_{7}=0.0845$, $w_{11}=0.0792$ \\
99 / 50 & 0 & $L_{11}$ & $w_{11}=0.5859$ \\
99 / 50 & 3 & $L_{0}$ & $w_{0}=0.9989$, $w_{11}\approx 10^{-6}$ \\ \bottomrule
\end{tabular*}}
\end{table}

These examples indicate that the router largely converges to the same intermediate layer while retaining occasional input-conditioned deviations. Thus, the learned policy is concentrated but not strictly equivalent to a fixed-layer rule.

\section{Spectral Distillation and Representation-Spectrum Consistency}
\label{app:spectral_discussion}

This section clarifies the representation-level spectral objective used by DLink. It separates the mathematical formulation from the interpretation boundary, because the objective operates on learned temporal-latent representations rather than on raw EEG frequency bands.

\subsection{Spectral Distillation Formulation}
\label{app:spectral_formulation}

Spectral distillation is applied to learned feature maps exposed to DLink. We use $f\in\mathbb{R}^{N\times d}$ as a notational view: $N$ denotes the ordered temporal-latent axis used for spectral alignment, and $d$ denotes all remaining feature indices. The Fourier transform is a one-dimensional transform $\mathcal{F}_N$ along the $N$ axis, computed independently over the remaining feature indices. It is not applied to raw EEG signals, but to learned representations before compression. Given student projection feature $f_S^p$ and routed teacher representation $f_T^*$, we compute
\begin{equation}
    \mathcal{F}_N(f_S^p)=M_S \odot e^{j\Phi_S},
    \qquad
    \mathcal{F}_N(f_T^*)=M_T \odot e^{j\Phi_T}.
\end{equation}
The magnitude consistency term is
\begin{equation}
    \mathcal{L}_{\mathrm{mag}}=\lVert M_S-M_T\rVert_F^2.
\end{equation}
To avoid phase discontinuity from $2\pi$ periodicity, phase is represented by sine and cosine components:
\begin{equation}
    \mathrm{Enc}(\Phi)=[\cos\Phi,\sin\Phi],
    \qquad
    \mathcal{L}_{\mathrm{phase}}
    =
    \lVert \mathrm{Enc}(\Phi_S)-\mathrm{Enc}(\Phi_T)\rVert_F^2.
\end{equation}
The spectral distillation loss is $\mathcal{L}_{\mathrm{kd}}=\mathcal{L}_{\mathrm{mag}}+\mathcal{L}_{\mathrm{phase}}$.

\subsection{Why Representation-Spectrum Consistency Helps Compact Transfer}
\label{app:frequency_consistency}

During compact transfer, projection and downsampling can alter learned feature spectra. Representation-spectrum consistency provides an auxiliary constraint that encourages the compact student feature to follow the spectral structure of the routed teacher representation before compression. This complements pointwise feature matching, which constrains Euclidean discrepancy but does not explicitly regularize the temporal-latent spectrum entering the compression block.

\subsection{Representation-Spectrum View}
\label{app:representation_spectrum_view}

The spectral bins used here describe variation across adjacent temporal latent units in a learned representation, not physiological EEG bands in Hz. This is meaningful because the temporal-latent axis preserves the order used by the feature layout entering compression, and the compression block reduces this latent resolution before classification.

Our motivation is representation-level rather than a strict signal-reconstruction claim. Anti-aliased CNNs show that ordered learned feature maps can benefit from frequency-aware treatment before resolution reduction~\cite{Shift_Invariant,CNN_aliasing}. Analogously, DLink constrains the pre-compression temporal-latent spectrum to follow the routed teacher spectrum. A smaller high-frequency tail after compression indicates smoother compact features and reduced representation-spectrum distortion.

\subsection{What We Do Not Claim}
\label{app:spectral_scope}

We do not claim that spectral distillation recovers physiological rhythms, identifies causal EEG frequency bands, or preserves task-relevant oscillatory activity in a neurophysiological sense. Our claim is limited to regularization for compact transfer: spectral distillation improves compact transfer and reduces distortion observed in learned representation spectra.

\subsection{Layer Score Definitions for Spectral-Concentration Guidance}
\label{app:spectral_energy_definitions}

The routing ablation compares three layer scores computed from a teacher feature $f_T^{(l)}$. Let
$A_l(\omega)=\lVert\mathcal{F}_N(f_T^{(l)})(\omega)\rVert_2$ denote the feature-index-aggregated spectral amplitude at temporal-latent representation-frequency bin $\omega \in \Omega$, and let $E_l(\omega)=A_l(\omega)^2$.
\begin{equation}
    \mathrm{MeanPower}_l
    =
    \frac{1}{|\Omega|}
    \sum_{\omega\in\Omega} E_l(\omega),
    \qquad
    \mathrm{MaxAmp}_l
    =
    \max_{\omega\in\Omega} A_l(\omega).
\end{equation}
The spectral-concentration score first forms a normalized energy distribution over representation-frequency bins:
\begin{equation}
    p_l(\omega)
    =
    \frac{E_l(\omega)}
    {\sum_{\omega'\in\Omega}E_l(\omega')+\epsilon}.
\end{equation}
We then reduce the distribution to a scalar layer score using spectral concentration:
\begin{equation}
    \mathrm{SpecConc}_l
    =
    \sum_{\omega\in\Omega} p_l(\omega)^2.
\end{equation}
In our experiments, this spectral-concentration score provides a more stable cue for routing than mean power or maximum amplitude.

\subsection{Focused Distillation Objective Comparison}
\label{app:distillation_objective}

We compare DLink with direct feature-level and logit-level distillation under the same student backbones. This focused comparison isolates the effect of routed representation-spectrum transfer from the use of a compact student alone.

\begin{table}[!htbp]
\centering
\caption{Comparison of distillation objectives on FACED using the same student backbones. Values are mean $\pm$ standard deviation over five seeds.}
\label{tab:distill_ablation}
\AppTableCaptionSep
\AppTightTableSetup
\resizebox{\textwidth}{!}{
\begin{tabular}{l|ccc|ccc}
\toprule
\multirow{2}{*}{\textbf{Objective}} & \multicolumn{3}{c|}{\textbf{Stu-S}} & \multicolumn{3}{c}{\textbf{Stu-M}} \\
& \textbf{ACC-B} & \textbf{Kappa} & \textbf{F1-W} & \textbf{ACC-B} & \textbf{Kappa} & \textbf{F1-W} \\ \midrule
Feature KD & 0.4239 {\scriptsize $\pm$ 0.0068} & 0.3486 {\scriptsize $\pm$ 0.0083} & 0.4217 {\scriptsize $\pm$ 0.0067} & 0.4858 {\scriptsize $\pm$ 0.0259} & 0.4191 {\scriptsize $\pm$ 0.0281} & 0.4889 {\scriptsize $\pm$ 0.0234} \\
Logit KD & 0.4125 {\scriptsize $\pm$ 0.0125} & 0.3367 {\scriptsize $\pm$ 0.0141} & 0.4117 {\scriptsize $\pm$ 0.0127} & 0.5122 {\scriptsize $\pm$ 0.0077} & 0.4483 {\scriptsize $\pm$ 0.0078} & 0.5117 {\scriptsize $\pm$ 0.0071} \\
\textbf{DLink (Ours)} & \textbf{0.4330 {\scriptsize $\pm$ 0.0132}} & \textbf{0.3548 {\scriptsize $\pm$ 0.0113}} & \textbf{0.4251 {\scriptsize $\pm$ 0.0107}} & \textbf{0.5221 {\scriptsize $\pm$ 0.0052}} & \textbf{0.4581 {\scriptsize $\pm$ 0.0066}} & \textbf{0.5202 {\scriptsize $\pm$ 0.0074}} \\ \bottomrule
\end{tabular}}
\end{table}

DLink gives the best FACED performance for both student scales, supporting the combination of layer routing and representation-spectrum consistency over direct feature or logit matching.

\section{Deployment-oriented Profiling Protocol}
\label{app:profiling_protocol}

\subsection{CPU-only Profiling Setup}
\label{app:cpu_profile_setup}

We evaluate pure forward latency under a conservative CPU-only setting: one CPU thread, batch size 1, no CUDA acceleration, 10 warm-up runs, and 100 timed runs. The reported values measure model forward computation only.

\subsection{Teacher, Student, and Router Timing}
\label{app:router_timing}
\label{app:router_profile}
\label{app:cpu_profile}

Table~\ref{tab:cpu_profile} reports the timing measurements used to support the deployment-oriented efficiency claim. Teacher and student latency are measured under the same CPU-only batch-1 setting, whereas the router is profiled separately as a training-time CUDA module.

\begin{table}[!htbp]
\centering
\caption{Forward-latency profiling for the fine-tuned teacher, compact student, and training-time router. Teacher and student timings use CPU-only batch-1 inference; the router timing reports its separate training-time CUDA forward cost.}
\label{tab:cpu_profile}
\label{tab:router_profile}
\AppTableCaptionSep
\AppTableSetup
\resizebox{\textwidth}{!}{
\begin{tabular}{llllllll}
\toprule
\textbf{Model/Module} & \textbf{Device} & \textbf{Threads/Batch} & \textbf{Input shape} & \textbf{Warm-up} & \textbf{Timed runs} & \textbf{Latency} & \textbf{Notes} \\ \midrule
Fine-tuned teacher & CPU & 1 / 1 & $(1,19,5,200)$ & 10 & 100 & 19.345 ms/sample & Pure forward latency \\
Compact student & CPU & 1 / 1 & $(1,19,5,200)$ & 10 & 100 & 3.653 ms/sample & $5.30\times$ speedup over teacher \\
Layer router & CUDA & batch 64 & $(64,38,5)$ & 10 & 1000 & 0.633 ms/batch & Training-time only; 0.04M params, 0.22M FLOPs \\ \bottomrule
\end{tabular}}
\end{table}

Because the router is discarded after training, deployment uses only the compact student timing in this table.

\subsection{Boundary of the Profiling Claim}
\label{app:profiling_boundary}

This profiling evaluates pure forward latency under controlled CPU-only or module-level CUDA settings. It is not validation on commercial BCI hardware, and it does not include EEG acquisition, preprocessing, communication, operating-system scheduling, power measurement, or end-to-end online streaming overhead.

\section{Additional Discussion and Limitations}
\label{app:limitations}

\textbf{Scope of deployment claims.}
DLink is evaluated through deployment-oriented profiling rather than real-device deployment. The profiling results support the efficiency of the compact inference path, but they should not be interpreted as validation on commercial BCI hardware.

\textbf{Scope of spectral claims.}
The spectral analysis supports regularization for compact transfer and reduced distortion in learned representation spectra. It does not establish a causal physiological interpretation of EEG frequency bands.

\textbf{Future work.}
Future work includes validation on real BCI hardware, online EEG streaming experiments, broader teacher EFMs and downstream tasks, larger subject populations, and energy or memory profiling.

\section{Broader Impacts}
\label{app:broader_impacts}

DLink may help reduce the inference cost of EEG foundation-model adaptation by transferring teacher knowledge into compact students. This can support more efficient research prototypes for EEG decoding and may lower the computational barrier for studying compact models in BCI, affective computing, and neurological assessment settings.

The main risk is reliability under realistic acquisition conditions. The experiments use existing benchmark datasets rather than online clinical or commercial-device recordings, and real-world EEG data may contain stronger sensor noise, subject variability, artifacts, and distribution shifts than those represented in the benchmarks. DLink should therefore not be used as a clinical decision tool or deployed in safety-critical BCI settings without additional validation on the target hardware, acquisition protocol, and user population.